\documentclass[11pt,a4paper]{article}
\usepackage[hyperref]{naaclhlt2019}
\usepackage{url}
\usepackage{times}
\usepackage{latexsym}
\usepackage{booktabs}
\usepackage{enumitem}
\usepackage{mathtools}
\usepackage{amssymb}

\usepackage{amsmath,amsfonts,bm}
\DeclareMathOperator*{\argmax}{arg\,max}

\newcommand{\sect}[1]{(\S\ref{#1})}

\usepackage{algorithm}
\usepackage[noend]{algpseudocode}
\usepackage{graphicx}
\usepackage{subcaption}
\usepackage{forest}
\forestset{
    nice empty nodes/.style={
        for tree={
            s sep=0.3em, 
            l sep=0.3em,
            inner ysep=0.5em, 
            inner xsep=0.45em,
            l=0,
            calign=midpoint,
            fit=tight,
            where n children=0{
               tier=word,
               minimum height=2.0em,
            }{},
            where n children=2{
               l-=1em,
            }{},
            parent anchor=south,
            child anchor=north,
            delay={if content={}{
                    inner sep=0pt,
                    edge path={\noexpand\path [\forestoption{edge}] 
                    			(!u.parent anchor) 
                               -- (.south)\forestoption{edge label};}
                }{}}
        },
    },
}

\newcommand\Tstrut{\rule{0pt}{2.6ex}}       
\newcommand\Bstrut{\rule[-0.9ex]{0pt}{0pt}} 

\aclfinalcopy 
\def\aclpaperid{862} 

\title{Unsupervised Latent Tree Induction with \\Deep Inside-Outside Recursive Autoencoders}

\author{Andrew Drozdov$^{1,*}$
  \\\And
  Pat Verga$^{1,*}$
  \\\And
  Mohit Yadav$^{1,}$\thanks{~\,Equal contribution, randomly ordered.}
  \\\And
  Mohit Iyyer$^{1}$
  \\\And
  Andrew McCallum$^{1}$ \\
  \AND\\[-5ex]
$^{1}$College of Information and Computer Sciences\\University of Massachusetts Amherst
\AND\\[-5ex]
    \texttt{\{adrozdov, pat, ymohit, miyyer, mccallum\}@cs.umass.edu}}

\date{}

\begin{document}
\maketitle

\begin{abstract}
  We introduce deep inside-outside recursive autoencoders (DIORA), a fully-unsupervised method for discovering syntax that simultaneously learns representations for constituents within the induced tree. Our approach predicts each word in an input sentence conditioned on the rest of the sentence and uses inside-outside dynamic programming to consider all possible binary trees over the sentence. At test time the CKY algorithm extracts the highest scoring parse. DIORA achieves a new state-of-the-art F1 in unsupervised binary constituency parsing (unlabeled) in two benchmark datasets, WSJ and MultiNLI.
\end{abstract}

\section{Introduction}
\label{sec:intro}

Syntactic parse trees are useful for downstream tasks such as relation extraction \citep{gamallo2012dependency}, semantic role labeling \citep{sutton2005joint,he2018jointly}, machine translation~\citep{aharoni2017towards,eriguchi2017learning,zaremoodi2018incorporating}, and text classification~\citep{li2006learning, tai2015improved}. Traditionally, supervised parsers trained on datasets such as the Penn Treebank~\citep{marcus1993building} are used to obtain syntactic trees. However, the treebanks used to train these supervised parsers are typically small and restricted to the newswire domain. Unfortunately, models trained on newswire treebanks tend to perform considerably worse when applied to new types of data, and creating new domain specific treebanks with syntactic annotations is expensive and time-consuming.

Motivated by the desire to address the limitations of supervised parsing and by the success of large-scale unsupervised modeling such as ELMo and BERT \cite{Peters:2018,devlin2018bert}, we propose a new deep learning method of unsupervised parser training that can extract both shallow parses (i.e., noun phrases or entities) and full syntactic trees from any domain or language automatically \emph{without requiring any labeled training data}.
In addition to producing parses, our model simultaneously builds representations for internal constituents that reflect syntactic and semantic regularities which can be leveraged by downstream tasks. 


Our model builds on existing work developing latent tree chart parsers \citep{socher2011semi, le2015forest, yogatama2016learning, maillard2017jointly, choi2018learning}. These methods produce representations for all internal nodes in the tree (cells in the chart), each generated as a soft weighting over all possible sub-trees \sect{sec:model}. Unfortunately, they still require sentence-level annotations during training, as they are all trained to optimize a downstream task, typically natural language inference.

\begin{figure}[t!]
\centering
    \scalebox{.65}{
    \hspace{-4mm}
    \includegraphics[height=0.26\linewidth]{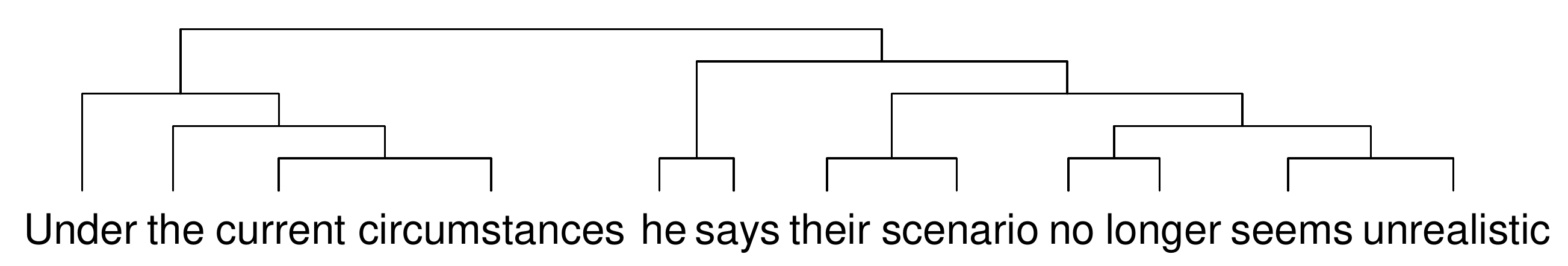}
}\\
	
	\caption{An unlabeled binary constituency parse from DIORA matching the ground truth.  \label{fig:trees_exact}
	}
	\vspace{-4mm}
\end{figure}

\begin{figure*}[t!]
\centering
\includegraphics[height=0.28\linewidth]{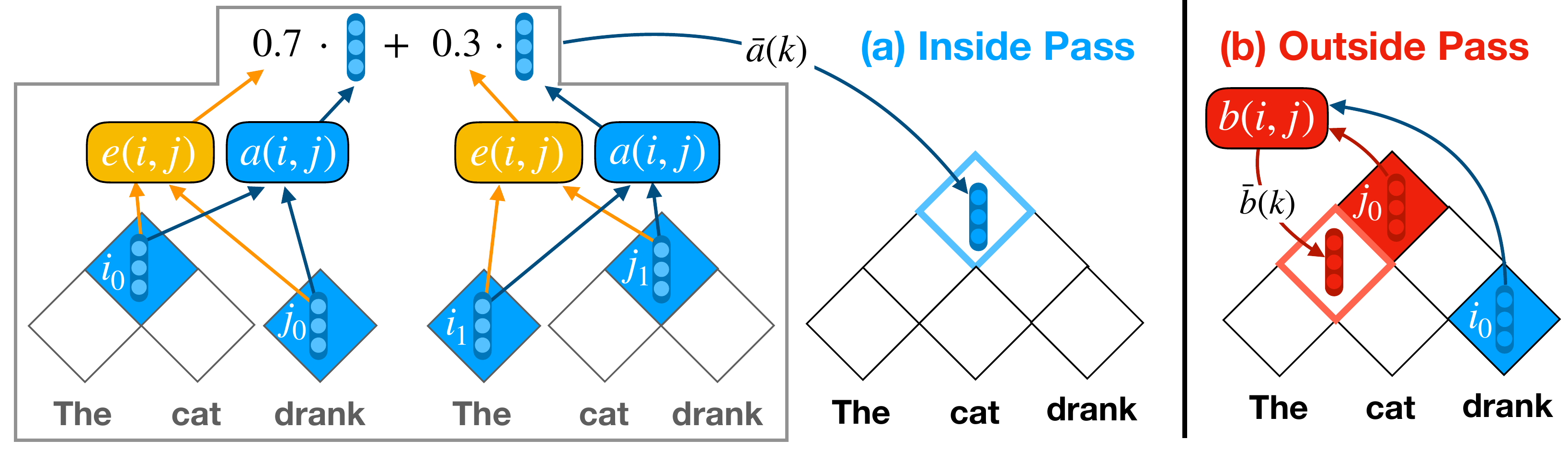}
\caption{The inside and outside pass of DIORA for the input `the cat drank'. a) The inside pass: The blue inside vector $\bar{a}(k)$ for the phrase `the cat drank' is a weighted average of the compositions for the two possible segmentations - ((the cat), drank) and (the, (cat drank)). The scalar weights come from a learned compatibility function.
b) The outside pass: The red outside vector $\bar{b}(k)$ for the phrase `the cat' is a function of the outside vector of its parent `the cat drank' and the inside vector of its sibling `drank'. }
\label{fig:model}
\end{figure*}

To address these limitations, we present deep inside-outside recursive autoencoders (DIORA) which enable unsupervised discovery and representation of constituents without requiring any supervised training data. DIORA incorporates the inside-outside algorithm~\citep{baker1979trainable,lari1990estimation} into a latent tree chart parser. The bottom-up inside step calculates a representation for all possible constituents within a binary tree over the input sentence. This step is equivalent to the forward-pass of previous latent tree chart parsers~\citep{maillard2017jointly}. These inside representations only encode the current subtree, ignoring all outside context. Thus, we perform an additional top-down outside calculation for each node in the tree, providing external context into the sub-tree representations in each chart cell. The model is then trained with the objective that the outside representations of the leaf cells should reconstruct the corresponding leaf input word, analogous to masked language model \citep{devlin2018bert} pre-training, except by using dynamic programming we predict every word from a completely unmasked context.
The single most likely tree can be recovered using the CKY algorithm and compatibility scores between constituents. Previous work either predict trees that are not well aligned with known treebanks \citep{yogatama2016learning, choi2018learning}, or has no mechanism for explicitly modeling phrases, requiring a complex procedure to extract syntactic structures \citep{shen2017neural}. 

To probe different properties of our model, we run experiments on unsupervised parsing, segment recall, and phrase representations. DIORA achieves multiple new state-of-the-art results for unsupervised constituency parsing (absolute improvements of \textbf{13.7\%}, \textbf{11.5\%}, and \textbf{7.8\%} on WSJ, WSJ-40, and MultiNLI), has a greater recall on more constituent types than a strong baseline, and produces meaningful phrase representations.

\section{DIORA: Deep Inside-Outside Recursive Autoencoders}
\label{sec:model}

Our goal is to design a model and unsupervised training procedure that learns structure from raw text. The design of DIORA is based on our hypothesis is that the most effective compression of a sentence will be derived from following the true syntactic structure of the underlying input. Our approach builds on previous latent tree chart parsers which are augmented with the inside-outside algorithm \citep{baker1979trainable,lari1990estimation} and trained to reproduce each input word from its outside context.
Based on our hypothesis, loosely inspired by the linguistic ``substitution principle'' \citep{frege}, the model will best reconstruct the input by discovering and exploiting syntactic regularities of the text.

The inside pass of our method recursively compresses the input sequence, at each step inputting the vector representations of the two children into a composition function \sect{ssec:inside} that outputs an inside vector representation of the parent. This process continues up to the root of the tree, eventually yielding a single vector representing the entire sentence (Figure \ref{fig:model}a). This is loosely analogous to the compression step of an autoencoder and equivalent to existing latent tree chart parsers forward pass \citep{maillard2017jointly}. Following this, we initiate the outside pass of our algorithm with a generic (root) representation that is learned as a separate parameter. As the outside step of the inside-outside algorithm (Figure \ref{fig:model}b), we unfold until finally producing representations of the leaf nodes. These leaves are then optimized to reconstruct the input sentence as done in an autoencoder-based deep neural network.

\subsection{Filling the Chart with Inside-Outside \label{ssec:fill_algorithm}}

Each inside representation is the root of a particularly sub-tree, and that representation is generated by considering only the descendant constituents within that sub-tree, ignoring any outside context. After the inside representations are calculated, we perform a top-down outside pass to compute outside representations. The outside representations are encoded by looking at only the context of a given sub-tree.  Once the chart is filled, each constituent $k$ (cell in the chart) is associated with an inside vector $\bar{a}(k)$, an outside vector $\bar{b}(k)$, inside compatibility score $\bar{e}(k)$ and outside compatibility score $\bar{f}(k)$.

The input to our model is a sentence $\boldsymbol{x}$ made up of $T$ tokens, $x_0, x_1, ..., x_{T-1}$. Each token $x_i$ has a corresponding pre-trained embedded vector $v_i$. 

\subsubsection{Inside Pass \label{ssec:inside}} 

For each pair of neighboring constituents $i$ and $j$~\footnote{The symbols $i$, $j$, and $k$ are identifiers of spans from the input $\boldsymbol{x}$. The symbol $i^*$ identifies a token from the set of negative examples $\{\boldsymbol{x}^*\}$.}, we compute a \textit{compatibility} score and a \textit{composition} vector. The score and vector that represent a particular span $k$ are computed using a soft weighting over all possible pairs of constituents, that together fully cover the span (we refer to this set of constituent pairs as $\{k\}$).

Vectors for spans of length 1 are initialized as a non-linear transformation~\footnote{This function shares its bias term $b$ with $\mathrm{Compose}_\alpha$, although $U_{\psi}$ is not tied to any other weights.} of the embedded input $v_i$, and the scores associated with these spans are set to $0$:

\vspace{-4mm}
\begin{align*}
    \begin{bmatrix}
        x \\
        o \\
        u \end{bmatrix} &= 
        \begin{bmatrix}
        \sigma \\
        \sigma \\
        \tanh \end{bmatrix}
        ( U_{\psi} v_k + b ) \\
\bar{a}(k) &= o + \tanh(x \odot u) \\
\bar{e}(k) &= 0
\end{align*}

Higher  levels  of  the  chart  are  computed  as a weighted summation of constituent pairs: 

\vspace{-2mm}
\begin{align*}
    \bar{a}(k) &= \sum\limits_{i,j \in \{k\}} e(i, j) ~a(i, j) \\
    \bar{e}(k) &= \sum\limits_{i,j \in \{k\}} e(i, j) ~\hat{e}(i, j)
\end{align*}

The compatibility function $\hat{e}$ is meant to produce a score for how likely a pair of neighboring cells are to be merged. We implement this as a bilinear function of the vectors from neighboring spans, using a learned parameter matrix $S$. We additionally add the individual scores from each two merging cells. Intuitively, these individual scores correspond to how likely each of the cells would exist in the final binary tree independently. The formula for the compatibility function (and its normalized form $e$) is defined as follows:

\vspace{-2mm}
\begin{align*}
    e(i, j) &= \frac{\exp(\hat{e}(i, j))}
    {\sum\limits_{\hat{i},\hat{j} \in \{k\}} \exp(\hat{e}(\hat{i}, \hat{j}))} \\
    \hat{e}(i, j) &= \phi(\bar{a}(i), \bar{a}(j); S_\alpha) + \bar{e}(i) + \bar{e}(j)
\end{align*}

Where the bilinear projection $\phi$ is defined as:

\vspace{-2mm}
\begin{align*}
    \phi(u, v; W) &= u^{\top} W v
\end{align*}

For the composition function $a$ we used either a $\mathrm{TreeLSTM}$ \citep{tai2015improved} or a 2-layer $\mathrm{MLP}$ (see Appendix \ref{sec:app_composition} for more precise definitons on both methods). In order for the remainder of equations to remain agnostic to the choice of composition function, we refer to the function as $\mathrm{Compose}$, which produces a hidden state vector $h$ and, in the case of $\mathrm{TreeLSTM}$, a cell state vector $c$, resulting in:

\vspace{-2mm}
\begin{align*}
    a(i, j) =
        \mathrm{Compose}_\alpha(\bar{a}(i), \bar{a}(j))
\end{align*}
        
\subsubsection{Outside Pass \label{ssec:outside}} 

The outside computation is similar to the inside pass (depicted in Figure \ref{fig:model}b).

The root node of the outside chart is learned as a bias. Descendant cells are predicted using a disambiguation over the possible outside contexts. Each component of the context consists of a sibling cell from the inside chart and a parent cell from the outside chart.

The function $f$ is analogous to the function $e$. It is normalized over constituent pairs $i,j$ for the span $k$, and is used to disambiguate among the many outside contexts. The function $b$ generates a phrase representation for the missing sibling cell. Equations for the outside computation follow:

\vspace{-2mm}
\begin{align*}
    \bar{b}(k) &= \sum\limits_{i,j \in \{k\}} f(i, j) ~b(i, j) \\
    \bar{f}(k) &= \sum\limits_{i,j \in \{k\}} f(i, j) ~\hat{f}(i, j) \\
    b(i, j) &= \mathrm{Compose}_{\beta}(\bar{a}(i), \bar{b}(j)) \\
    \hat{f}(i, j) &= \phi(\bar{a}(i), \bar{b}(j); S_\beta) + \bar{e}(i) + \bar{f}(j)
\end{align*}

 In the majority of our experiments, the $\mathrm{Compose}$ used in $b$ shares parameters with $a$ used in the inside pass, as do the compatibility functions $\hat{e}$ and $\hat{f}$ (see \S\ref{sec:modeling_choices} for results on the effects of parameter sharing).

\subsection{Training Objective \label{ssec:objective}}

To train our model we use an autoencoder-like language modeling objective. In a standard autoencoder, the entire input $\boldsymbol{x}$ is compressed into a single lower dimensional representation. This representation, $\boldsymbol{z}$, is then decompressed and trained to reconstruct $\boldsymbol{x}$. In our model, we never condition the reconstruction of $\boldsymbol{x}$ on a single $\boldsymbol{z}$ because the root's outside representation is initialized with a bias rather than the root's own inside vector. Instead, we reconstruct $\boldsymbol{x}$ conditioned on the many sub-tree roots, each of which is only a compression of a \textit{subset} of the input.

To approximate this reconstruction we use a max-margin loss considering a set $\{\boldsymbol{x^*}\}$ of $N$ negative examples that are sampled according to their frequency from the vocabulary (further details in Appendix \ref{sec:app_training}). The terminal outside vector $\bar{b}(i)$ is trained to predict its original input $v_i$.

The per-instance loss function is described in Equation \ref{eq:objective}:

\vspace{-2mm}
\begin{align}
    L_{\boldsymbol{x}} =
        \sum\limits_{i=0}^{T-1} 
        \sum\limits_{i^*=0}^{N-1}
        \max ( 0, 1
            &- \bar{b}(i) \cdot \bar{a}(i) \notag\\
            &+ \bar{b}(i) \cdot \bar{a}(i^*)
            )
        \label{eq:objective}
\end{align}

The max-margin loss does not provide a gradient if the predicted vector is closer to its ground truth than the negative example by a margin greater than $1$. For that reason, we also experimented with an objective based on cross-entropy, described in Equation \ref{eq:objective_xent}:

\vspace{-2mm}
\begin{align}
    Z^* &= \sum\limits_{i^*=0}^{N-1}
        \exp(\bar{b}(i) \cdot \bar{a}(i^*)) \nonumber \\
    L_{\boldsymbol{x}} &=
        - \sum\limits_{i=0}^{T-1} 
        \log
        \frac{\exp(\bar{b}(i) \cdot \bar{a}(i))}
        {
        \exp(\bar{b}(i) \cdot \bar{a}(i)) + Z^*} 
        \label{eq:objective_xent}
\end{align}

\subsection{DIORA CKY Parsing  \label{ssec:CKY}}

To obtain a parse with DIORA, we populate an inside and outside chart using the input sentence. We can extract the maximum scoring parse based on our single grammar rule using the CKY procedure \citep{kasami1966efficient,younger1967recognition}. The steps for this procedure are described in Algorithm \ref{alg:cky} and its runtime complexity in Appendix \ref{sec:app_runtime}.

\begin{algorithm}
  \caption{Parsing with DIORA}
  \label{alg:cky}
  \begin{algorithmic}[1]
      \Procedure{CKY}{$\mathrm{chart}$}
      \Statex \hskip1.5em \textit{Initialize terminal values.}
      \For{\textbf{each} $k \in \mathrm{chart} \mid \textproc{size}(k) = 1$}
        \State $x_k \gets 0$
      \EndFor
      \Statex \hskip1.5em \textit{Calculate a maximum score for each span,}
      \Statex \hskip1.5em \textit{and record a backpointer.}
      \For{\textbf{each} $k \in \mathrm{chart}$}
        \State $x_k \gets \max\limits_{i,j \in \{k\}} [x_i + x_j + e(i,j)]$
        \State $\pi^i_k,\pi^j_k \gets \argmax\limits_{i,j \in \{k\}} [x_i + x_j + e(i,j)]$
      \EndFor
      \Statex \hskip1.5em \textit{Backtrack to get the maximal tree.}
      \Procedure{Backtrack}{$k$}
        \If {\textproc{size}$(k) = 1$}
            \State \textbf{return} k
        \EndIf
        \State $i \gets$ \textproc{Backtrack}($\pi_k^i$)
        \State $j \gets$ \textproc{Backtrack}($\pi_k^j$)
        \State \textbf{return} (i, j)
      \EndProcedure
      \State \textbf{return} \textproc{Backtrack}($k \gets \mathrm{root}$)
    \EndProcedure
  \end{algorithmic}
\end{algorithm}

\section{Experiments}
\label{sec:experiments}

To evaluate the effectiveness of DIORA, we run experiments on unsupervised parsing, unsupervised segment recall, and phrase similarity. The model has been implemented in PyTorch \citep{pytorch} and the code is published online.\footnote{\url{https://github.com/iesl/diora}} For training details, see Appendix \ref{sec:app_training}.

\subsection{Unsupervised Parsing}
\label{ssec:unsupervised_parsing}

We first evaluate how well our model predicts a full unlabeled constituency parse. We look at two data sets used in prior work \citep{htut2018emnlp}, The Wall Street Journal (WSJ) section of Penn Treebank \citep{marcus1993building}, and the automatic parses from MultiNLI \citep{N18-1101}. WSJ has gold human-annotated parses and MultiNLI contains automatic parses derived from a supervised parser \citep{manning2014stanford}. 

In addition to PRPN \cite{shen2017neural},\footnote{We consider the PRPN models using LM stopping criteria, which outperformed UP.} we compare our model to deterministically constructed left branching, right branching, balanced, and random trees. We also compare to ON-LSTM \cite{shen2018ordered}, an extension of the PRPN model, RL-SPINN \citep{yogatama2016learning}, an unsupervised shift-reduce parser, and ST-Gumbel \citep{choi2018learning}, an unsupervised chart parser. The latter two of these models are trained to predict the downstream task of natural language inference (NLI).

\subsubsection{Binarized WSJ and MultiNLI results\label{sec:wsj_nli}}
For the full WSJ test set and MultiNLI datasets we follow the experimental setup of previous work \citep{williams2018tacl}. We binarize target trees using Stanford CoreNLP~\citep{manning2014stanford} and do not remove punctuation (experiments in \S\ref{ssec:wsj_10_40} do remove punctuation).

Latent tree models have been shown to perform particularly poorly on attachments at the beginning and end of the sequence \citep{williams2018tacl}. To address this, we incorporate a post-processing heuristic (denoted as $\mathrm{+PP}$ in result tables)\footnote{We did not have access to predictions or an implementation of the concurrent ON-LSTM model and therefore could not apply the ${\mathrm{+PP}}$ heuristic.}. This heuristic simply attaches trailing punctuation to the root of the tree, regardless of its predicted attachment.

In Table \ref{tab:wsj_results}, we see that DIORA$^{\mathrm{+PP}}$ achieves the highest average and maximum F1 from five random restarts. This model achieves a mean F1 7 points higher than ON-LSTM and an increase of over 6.5 max F1 points. We also see that DIORA exhibits much less variance between random seeds than ON-LSTM. Additionally, we find that PRPN-UP and DIORA benefit much more from the ${\mathrm{+PP}}$ heuristic than PRPN-LM. This is consistent with qualitative analysis showing that DIORA and PRPN-UP incorrectly attach trailing punctuation much more often than PRPN-LM.

On the MultiNLI dataset, PRPN-LM is the top performing model without using the $\mathrm{+PP}$ heuristic while DIORA matches PRPN-UP (Table \ref{tab:nli_results}. Using the heuristic, DIORA greatly surpasses both variants of PRPN. However, it is worth noting that this is not a gold standard evaluation and instead evaluates a model's ability to replicate the output of a trained parser \citep{manning2014stanford}. A second caveat is that SNLI \cite{snli2015emnlp} and MultiNLI contain several non-newswire domains. Syntactic parsers often suffer significant performance drops when predicting outside of the newswire domain that the models were trained on. 

\begin{table}[ht!]
\centering
{
\renewcommand{\arraystretch}{1.0} 
\begin{tabular}{l|lcc}
\toprule
 \textbf{Model} & \textbf{$\textrm{F1}_\mu$} & \textbf{$\textrm{F1}_{max}$} & \textbf{$\delta$}  \\
 \midrule
LB                              & 13.1            & 13.1        & 12.4\Tstrut   \\
RB                              &  16.5           &  16.5       & 12.4          \\
Random                          & 21.4            & 21.4        & 5.3           \\
Balanced                        & 21.3                & 21.3        & 4.6           \\
RL-SPINN$\dagger$               & 13.2               &  13.2        & -             \\
ST-Gumbel \small{- GRU}$\dagger$     &  22.8  \scriptsize${\pm 1.6}$     &  25.0       & -\Bstrut \\ 
\midrule
PRPN-UP                           &  38.3 \scriptsize${\pm  0.5}$   &  39.8        & 5.9\Tstrut \\
PRPN-LM                           &  35.0  \scriptsize${\pm 5.4}$    &  42.8       & 6.2           \\
ON-LSTM                           &  47.7  \scriptsize${\pm 1.5}$    &  49.4       & 5.6           \\
DIORA                             & 48.9  \scriptsize${\pm 0.5}$     & 49.6        & 8.0\Bstrut \\ 
\midrule
PRPN-UP$^{\mathrm{+PP}}$          &  -                  &  45.2     & 6.7\Tstrut \\
PRPN-LM$^{\mathrm{+PP}}$          &  -                  &  42.4     & 6.3           \\
DIORA$^{\mathrm{+PP}}$           & \textbf{55.7} \scriptsize${\pm 0.4}$   & \textbf{56.2}     & 8.5 \\
\bottomrule
\end{tabular}
}
\normalsize
\caption{\textbf{Full WSJ (test set) unsupervised unlabeled binary constituency parsing including punctuation.} $\dagger$ indicates trained to optimize NLI task. Mean and max are calculated over five random restarts. PRPN F1 was calculated using the parse trees and results provided by \citet{htut2018emnlp}. The depth ($\delta$) is the average tree height.  $\mathrm{+PP}$ refers to post-processing heuristic that attaches trailing punctuation to the root of the tree. The top F1 value in each column is bolded.
}
\label{tab:wsj_results}
\end{table}

\begin{table}[ht!]
\centering
{
\renewcommand{\arraystretch}{1.0} 
\begin{tabular}{l|lcc}
\toprule
 \textbf{Model} & \textbf{$\textrm{F1}_{median}$} & \textbf{$\textrm{F1}_{max}$} & \textbf{$\delta$}  \\
 \midrule
Random                          & 27.0            & 27.0        & 4.4          \\
Balanced                        & 21.3                & 21.3        & 3.9           \\
\midrule
PRPN-UP                           &  48.6    &  -        & 4.9\Tstrut \\
PRPN-LM                           &  50.4     &  -       & 5.1          \\
DIORA                             & 51.2      & 53.3        & 6.4 \Bstrut \\ 
\midrule
PRPN-UP$^{\mathrm{+PP}}$          &  -                  &  54.8     & 5.2 \Tstrut \\
PRPN-LM$^{\mathrm{+PP}}$          &  -                  &  50.4    & 5.1           \\
DIORA$^{\mathrm{+PP}}$           & 59.0   & \textbf{59.1}     & 6.7 \\
\bottomrule
\end{tabular}
}
\normalsize
\caption{\textbf{NLI unsupervised unlabeled binary constituency parsing comparing to CoreNLP predicted parses.} PRPN F1 was calculated using the parse trees and results provided by \citet{htut2018emnlp}. F1 median and max are calculated over five random seeds and the top F1 value in each column is bolded. Note that we use median rather than mean in order to compare with previous work.
}
\label{tab:nli_results}
\vspace{-4mm}
\end{table}

\subsubsection{WSJ-10 and WSJ-40 results}
\label{ssec:wsj_10_40}

We also compare our models to two subsets of the WSJ dataset that were used in previous unsupervised parsing evaluations. WSJ-10 and WSJ-40 contain sentences up to length 10 and 40 respectively after punctuation removal. We do not binarize either of these two splits in order to compare to previous work (see Appendix \ref{sec:datasets} for more details on WSJ split differences). Not binarizing the target trees sets an upper-bound on the performance of our models, denoted as UB in Table \ref{tab:wsj10_results}.

We compare against previous notable models for this task: CCM \citep{Klein2002AGC} uses the EM algorithm to learn probable nested bracketings over a sentence using gold or induced part-of-speech tags, and PRLG \citep{Ponvert2011SimpleUG} performs constituent parsing through consecutive rounds of sentence chunking. 

In Table \ref{tab:wsj10_results}, we see that DIORA outperforms the previous state of the art for WSJ-40, PRLG, in max F1.  The WSJ-10 split has been difficult for latent tree parsers such as DIORA, PRPN, and ON-LSTM, none of which (including our model) are able to improve upon previous non-neural methods. However, when we compare trends between WSJ-10 and WSJ-40, we see that DIORA does a better job at extending to longer sequences.

\subsection{Unsupervised Phrase Segmentation
\label{ssec:unsupervised_phrase_segmentation}}

In many scenarios, one is only concerned with extracting particular constituent phrases rather than a full parse. Common use cases would be identifying entities, noun phrases, or verb phrases for downstream analysis. To get an idea of how well our model can perform on phrase segmentation, we consider the maximum recall of spans in our predicted parse tree. We leave methods for cutting the tree to future work and instead consider the maximum recall of our model which serves as an upper bound on its performance. Recall here is the percentage of labeled constituents that appear in our predicted tree relative to the total number of constituents in the gold tree. These scores are separated by type and presented in Table \ref{tab:unsupervised_segmentation}.

In Table \ref{tab:unsupervised_segmentation} we see the breakdown of constituent recall across the 10 most common types. DIORA achieves the highest recall across the most types and is the only model to perform effectively on verb-phrases. Interestingly, DIORA performs worse than PRPN-LM at prepositional phrases.

\subsection{Phrase Similarity}
\label{ssec:phrase}
One of the goals of DIORA is to learn meaningful representations for spans of text. Most language modeling methods focus only on explicitly modeling token representations and rely on ad-hoc post-processing to generate representations for longer spans, typically relying on simple arithmetic functions of the individual tokens.

\begin{table}[ht!]
    \centering
    \setlength\tabcolsep{4pt}
    \resizebox{\columnwidth}{!}{%
    \begin{tabular}{l|lc|lc}
        \toprule
           &  \multicolumn{2}{c}{ \textbf{WSJ-10}} &  \multicolumn{2}{c}{ \textbf{WSJ-40}} \\
\textbf{Model} &  \textbf{$\textrm{F1}_\mu$} & \textbf{$\textrm{F1}_{max}$} &  \textbf{$\textrm{F1}_\mu$} & \textbf{$\textrm{F1}_{max}$}\\
\midrule
UB               & 87.8 & 87.8           &  85.7 & 85.7 \\
LB              &  28.7 & 28.7            & 12.0 & 12.0          \\
RB              &  61.7 & 61.7          & 40.7  &  40.7         \\
\midrule 
$\textrm{CCM}\dagger$  & -  &  63.2     & - & -         \\
$\textrm{CCM}_{\small{gold}}\dagger$      & -  &  71.9    & -    & 33.7              \\
PRLG $\dagger$          &  - & \textbf{72.1}   & -   & 54.6    \\
\midrule
$\textrm{PRPN}_{NLI}$             & 66.3 \scriptsize{$\pm  0.8$}     &  68.5   & -   & -        \\
PRPN$\ddagger$                               & 70.5 \scriptsize{$\pm 0.4$}     &  71.3   & -   & 52.4        \\
ON-LSTM$\ddagger$                            & 65.1 \scriptsize{$\pm 1.7$} &  66.8   & - & -     \\
DIORA                               &  67.7 \scriptsize{$\pm 0.7$}     & 68.5 & 60.6 \scriptsize{$\pm 0.2$}  & \textbf{60.9}  \\
\bottomrule
\end{tabular}
}
\caption{\textbf{WSJ-10 and WSJ-40 unsupervised non-binary unlabeled constituency parsing with punctuation removed.} $\dagger$ indicates that the model predicts a full, non-binary parse with additional resources. $\ddagger$ indicates model was trained on WSJ data and $\textrm{PRPN}_{NLI}$ was trained on MultiNLI data. CCM uses predicted POS tags while $\textrm{CCM}_{\small{gold}}$ uses gold POS tags. PRPN F1 was calculated using the parse trees and results provided by \citet{htut2018emnlp}. LB and RB are the left and right-branching baselines. UB is the upper bound attainable by a model that produces binary trees.}
\label{tab:wsj10_results}
\end{table}

To evaluate our model's learned phrase representations, we look at the similarity between spans of the same type within labeled phrase datasets. We look at two datasets. CoNLL 2000 \citep{tjong2000introduction} is a shallow parsing dataset containing spans of noun phrases, verb phrases, etc. CoNLL 2012 \citep{pradhan2012conll} is a named entity dataset containing 19 different entity types. 

\begin{table}[h] 
\centering
{\renewcommand{\arraystretch}{1.2} 
\begin{tabular}{lr|ccc} 
\toprule
Label &  Count & DIORA & P-UP & P-LM \\
 \midrule
NP      & 297,872   &  \textbf{0.767}    &  0.687    & 0.598 \\
VP      & 168,605   &  \textbf{0.628}    &  0.393    & 0.316 \\
PP      & 116,338   &  0.595    &  0.497    & \textbf{0.602} \\
S       & 87,714    &  \textbf{0.798}    &  0.639    & 0.657 \\
SBAR    & 24,743    &  \textbf{0.613}    &  0.403    & 0.554 \\
ADJP    & 12,263    &  \textbf{0.604}    &  0.342    & 0.360 \\
QP      & 11,441    &  \textbf{0.801}    &  0.336    & 0.545 \\
ADVP    & 5,817     &  \textbf{0.693}    &  0.392    & 0.500 \\
PRN     & 2,971     &  \textbf{0.546}    &  0.127    & 0.144 \\
SINV    & 2,563     &  0.926    &  0.904    & \textbf{0.932} \\
\bottomrule
\end{tabular}
}
\caption{\textbf{Segment recall from WSJ separated by phrase type.} The 10 most frequent phrase types are shown above, and the highest value in each row is bolded. P-UP=PRNP-UP, P-LM=PRPN-LM}
\label{tab:unsupervised_segmentation}
\end{table}

For each of the labeled spans with length greater than one, we first generate its phrase representation. We then calculate its cosine similarity to all other labeled spans. We then calculate if the label for that query span matches the labels for each of the $K$ most similar other spans in the dataset. In Table \ref{tab:phrase_sim_chunks} we report precision@$K$ for both datasets and various values of $K$.

The first baseline we compare against produces phrase representations from averaging context-insensitive (CI) ELMo vectors of individual tokens with the span. The second uses sentence-insensitive (SI) ELMo vectors, running the full ELMo over only the relevant tokens and ignoring the rest of the sentence. We also look at ELMo's output when given the entire sentence. When analyzing our baselines that run the full ELMo, we follow the procedure described in \citep{peters2018dissecting} and represent phrases as a function of its first and last hidden state. We extract these states from the final ELMo layer (3rd BiLSTM) as these consistently gave the best performance among other options. For DIORA, we use the concatenation of the inside and outside representations ($[\bar{a}; \bar{b}]$).

For CoNLL 2000, we find that our model outperforms all baselines for all values of $K$. This demonstrates DIORA's ability to capture and represent syntactic information within phrases. For CoNLL 2012, we find that DIORA outperforms both ELMo$_{CI}$ and ELMo$_{SI}$ while ELMo performs best overall. ELMo$_{CI}$ is surprisingly effective on this dataset even though it performed more poorly on CoNLL 2000. These results indicate that DIORA is capturing syntax quite well, but still has room to improve on  more fine-grained semantic representations. 

\begin{table*}[ht]
\centering
{\renewcommand{\arraystretch}{1.2} 
\begin{tabular}{l|c|ccc|ccc}
\toprule
 &  & \multicolumn{3}{c|}{ \textbf{CoNLL 2000}}  & \multicolumn{3}{c}{ \textbf{CoNLL 2012}} \\
 
\textbf{Model} & Dim  & P@1 & P@10  & P@100  & P@1 & P@10 & P@100  \\
 \midrule
Random & 800 & 0.684 & 0.683 & 0.680   & 0.137 & 0.133 & 0.135  \\ 
$\textrm{ELMo}_{\small{CI}}$ & 1024 & 0.962 & 0.955 & 0.957   & 0.708 & 0.643 & 0.544  \\
$\textrm{ELMo}_{\small{SI}}$ & 4096 & 0.970 & 0.964 & 0.955   & 0.660 & 0.624 & 0.533  \\ 
ELMo & 4096 & 0.987 & 0.983 & 0.974   & \textbf{0.896} & \textbf{0.847} & \textbf{0.716}  \\ 
\midrule
$\textrm{DIORA}_{\small{In/Out}}$  & 800 & \textbf{0.990} & \textbf{0.985} & \textbf{0.979} & 0.860 & 0.796 & 0.646  \\
\bottomrule
\end{tabular}
}
\caption{\textbf{P@1, P@10, and P@100 for labeled chunks from CoNLL-2000 and CoNLL 2012 datasets.} For all metrics, higher is better. The top value in each column is bolded. Diora uses the concatenation of the inside and outside vector at each cell which performed better than either in isolation.}
\label{tab:phrase_sim_chunks}
\end{table*}

\subsection{Impact of Modeling Choices \label{sec:modeling_choices}}
To test the impact of our modeling choices, we compared the performance of two different losses and four different composition functions on the full WSJ validation set. The losses were covered in Equations \ref{eq:objective} (Margin) and \ref{eq:objective_xent} (Softmax). The two primary methods of composition we considered were TreeLSTM \cite{tai2015improved} and MLP (a 2-hidden layer neural network). In addition, we experimented with a simple \textit{kernel} of the MLP input $[x; y; x \odot y; x - y]$ and with a setting where both the inside and outside parameters are \textit{shared}.

The results are shown in Table \ref{tab:ablation}. We see that MLP composition consistently performs better than with TreeLSTM, that MLP benefits from the Softmax loss, and that the best performance comes from sharing parameters. All other experimental results use this highly performant setting unless otherwise specified.

\newpage

\begin{figure}[!ht]
\centering
\resizebox{\columnwidth}{!}{%
    \includegraphics[width=\linewidth]{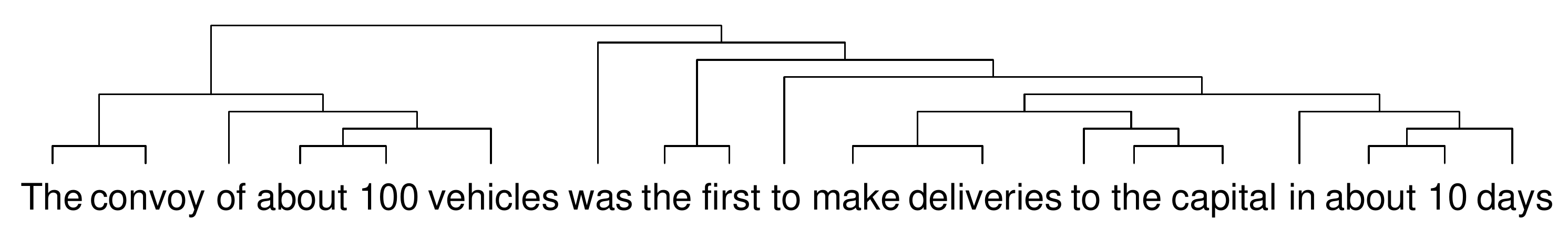}
}\\
\resizebox{\columnwidth}{!}{%
    \includegraphics[width=\linewidth]{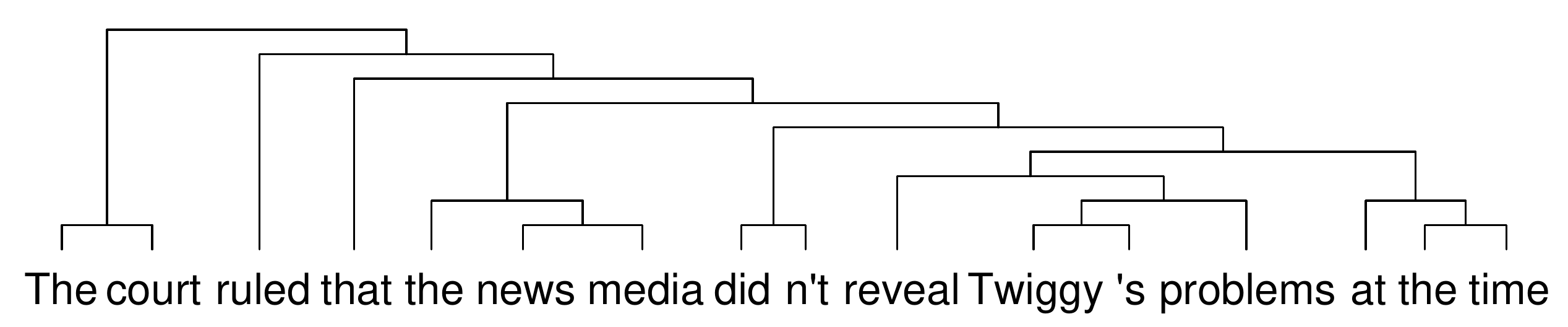}
}\\
\vspace{-1mm}
	\caption{DIORA can match the ground truth exactly. \label{fig:trees_exactgt}
	}
	\vspace{-6mm}
\end{figure}

\begin{figure}[h]
\centering
\resizebox{\columnwidth}{!}{%
    \includegraphics[width=\linewidth]{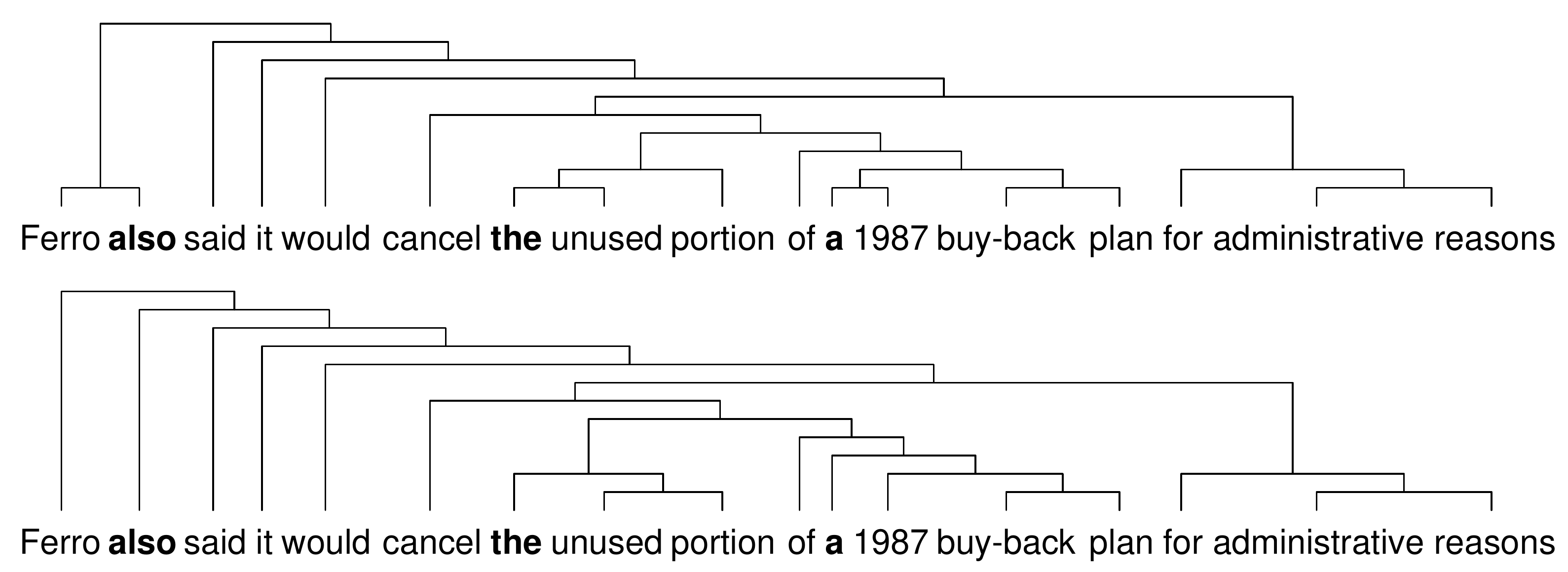}
}\\
\resizebox{\columnwidth}{!}{%
    \includegraphics[width=\linewidth]{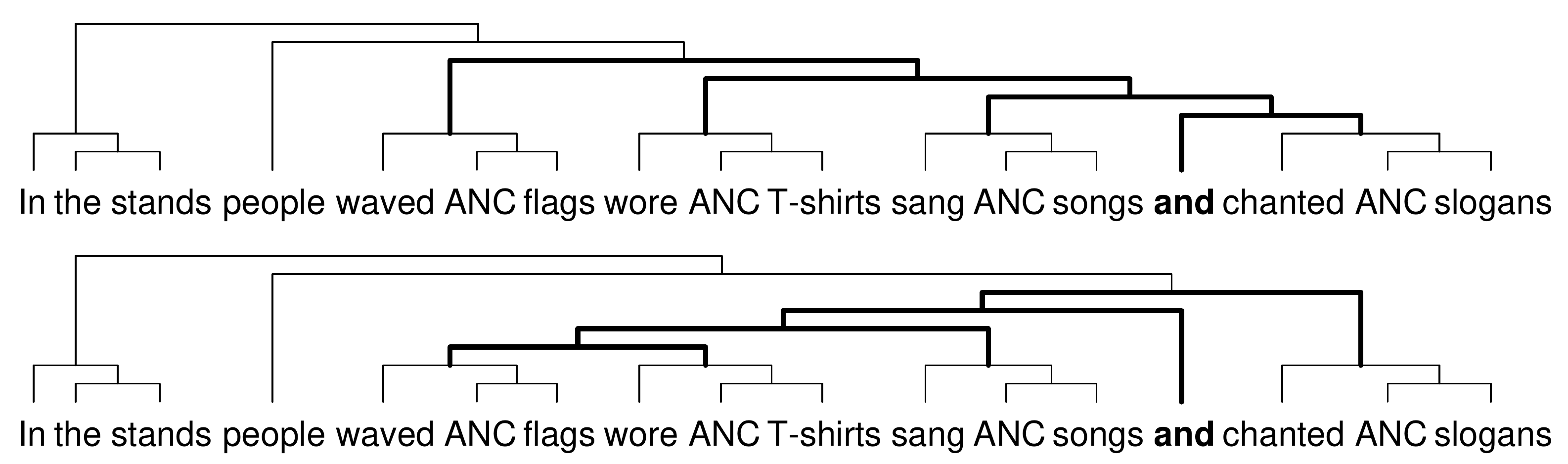}
}\\
\vspace{-1mm}
	\caption{At times, DIORA exhibits contrary behavior to the ground truth inevitably leading to some error. DIORA's output is shown above the ground truth.\footnotemark  \label{fig:trees_close}
	}
	\vspace{-6mm}
\end{figure}

\begin{figure}[!h]
\centering
\resizebox{\columnwidth}{!}{%
    \includegraphics[width=\linewidth]{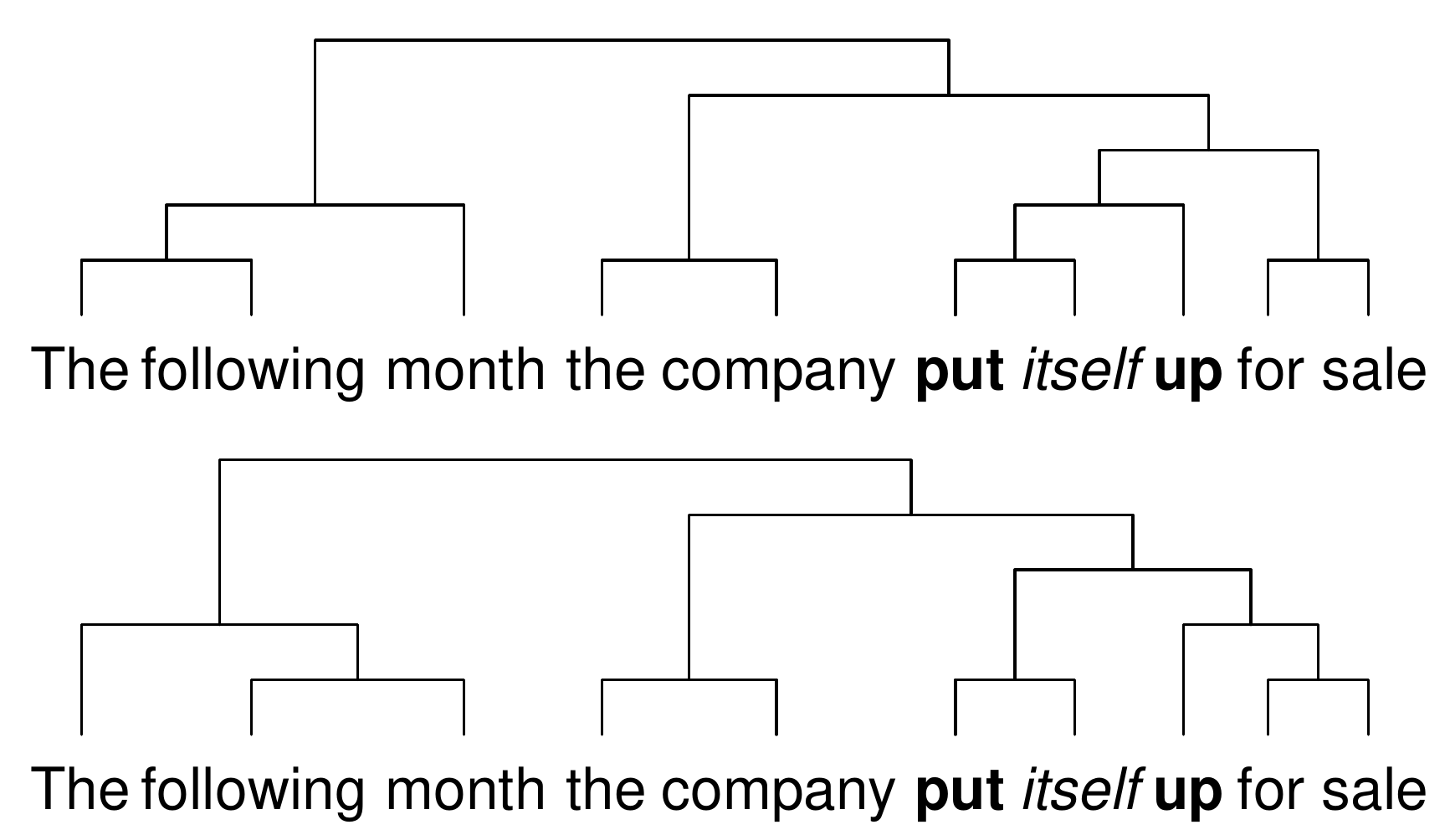}
}\\
\resizebox{\columnwidth}{!}{%
    \includegraphics[width=\linewidth]{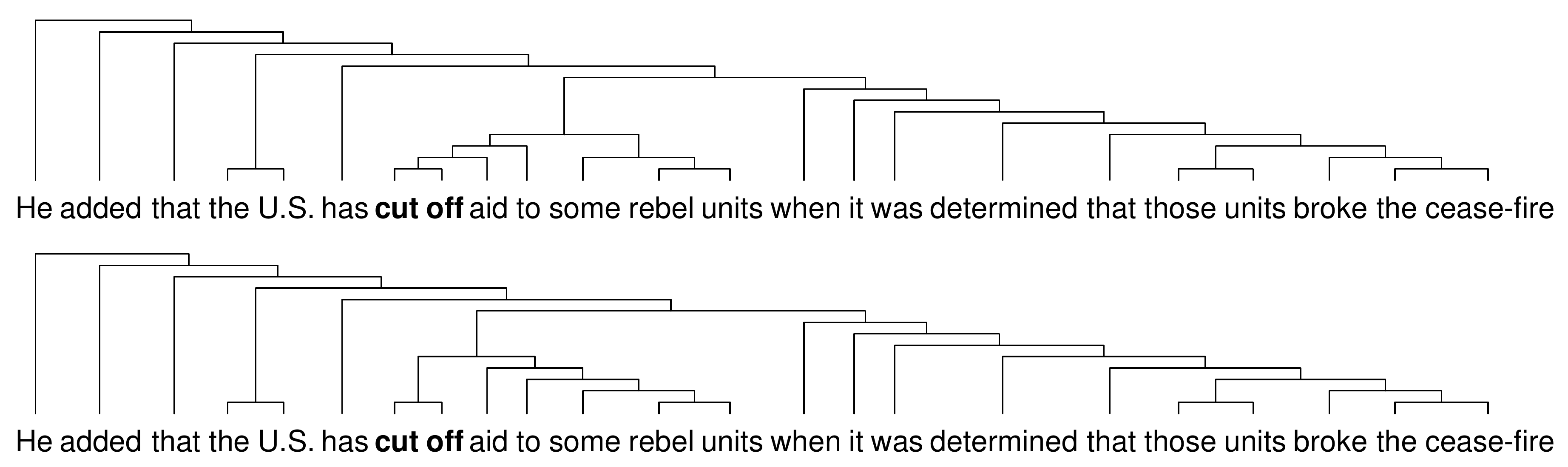}
}\\
\resizebox{\columnwidth}{!}{%
    \includegraphics[width=\linewidth]{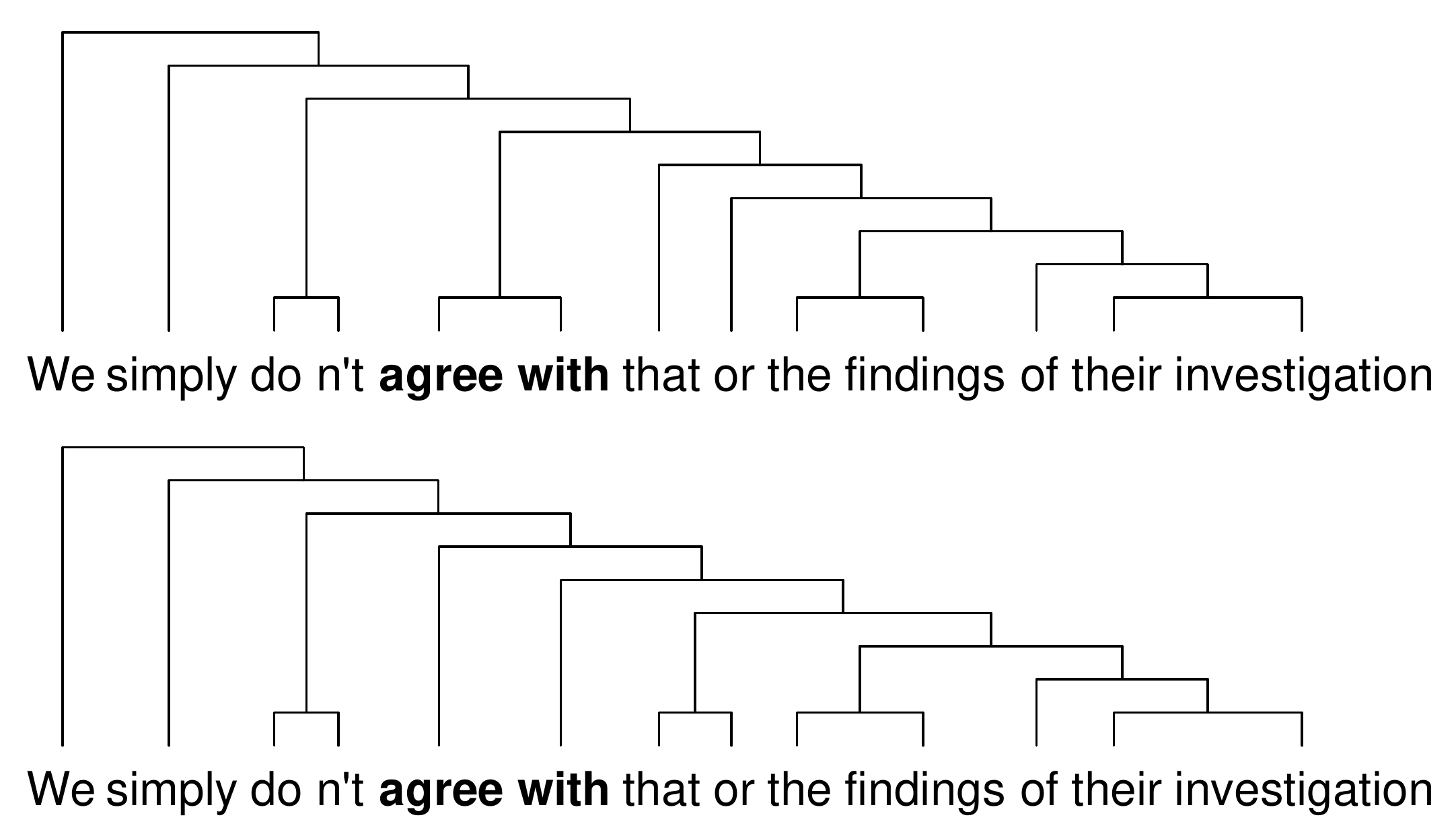}
}\\
\vspace{-1mm}
	\caption{DIORA often groups verbs and particles (top), sometimes exactly as the ground truth (middle). Occasionally, errors are particle-like (bottom). DIORA's output is shown above the ground truth.\footnotemark[6] \label{fig:trees_particle}
	}
	\vspace{-4mm}
\end{figure}

\footnotetext{Ground truth parses are binarized unless otherwise specified. All examples of DIORA parses are already binary. Some punctuation has been removed for easier readability.}

\subsection{Qualitative Results\label{ssec:qual_results}}

Looking at our model's output, we see that some trees are an exact replication of the binarized ground truth (Fig. \ref{fig:trees_exactgt}), or very close (Fig. \ref{fig:trees_close}). For future work we intend to explore common patterns in DIORA's learned structure, although some patterns are already recognizable, such as the affinity to group particles and verbs (Fig. \ref{fig:trees_particle}).


\section{Related Work} 
\label{sec:relatwed_work}

\textbf{Latent Tree Learning} A brief survey of neural latent tree learning models was covered in \cite{williams2018tacl}. The first positive result for neural latent tree parsing was shown in \cite{htut2018emnlp}, which used a language modeling objective. The model in \cite{liu2018emnlp} uses an inside chart and an outside procedure to calculate marginal probabilities in order to align spans between sentences in entailment.

\begin{table}[h] 
\centering
{\renewcommand{\arraystretch}{1.2} 
\begin{tabular}{cc|cc} 
\toprule
 &  &  \multicolumn{2}{c}{\textbf{$\textrm{F1}_{\mu}$}} \\
\textbf{Composition} & \textbf{Loss} & $\varnothing$ & ${\mathrm{+PP}}$ \\
 \midrule
TreeLSTM & Margin & 49.9 & 53.1  \\
TreeLSTM & Softmax  & 52.0 & 52.9  \\
MLP & Margin  & 49.7 & 54.4  \\
MLP & Softmax  & 52.6 & 55.5  \\
MLP$_{\mathrm{Kernel}}$ & Softmax  & 51.8 & 54.8  \\
MLP$_{\mathrm{Shared}}$ & Softmax  & 50.8 & 56.7  \\
\bottomrule
\end{tabular}
}
\caption{\textbf{F1 for different model variants on the binary WSJ validation set with included punctuation.} The binary trees are as-is ($\varnothing$) or modified according to the post-processing heuristic ($+PP$). The mean F1 is shown across three random seeds.}
\label{tab:ablation}
\end{table}

\textbf{Neural Inside-Outside Parsers} The Inside-Outside Recursive Neural Network (IORNN) \citep{le2014emnlp} is closest to ours. It is a graph-based dependency parser that uses beam search and can reliably find accurate parses when retaining a $k$-best list. In contrast, our model produces the most likely parse given the learned compatibility of the constituents. The Neural CRF Parser \citep{durrett2015acl}, similar to DIORA, performs exact inference on the structure of a sentence, although requires a set of grammar rules and labeled parse trees during training. DIORA, like \citet{liu2018emnlp}, has a single grammar rule that applies to any pair of constituents and does not use structural supervision.

\textbf{Learning from Raw Text}
Unsupervised learning of syntactic structure has been an active research area \citep{brill1990deducing}, including for unsupervised segmentation  \citep{ando2000mostly,goldwater2009bayesian,Ponvert2011SimpleUG} and unsupervised dependency parsing \cite{spitkovsky2013breaking}. Some models exploit the availability of parallel corpora in multiple languages \citep{Das2011UnsupervisedPT,Cohen2011UnsupervisedSP}. Others have shown that dependency parsing can be used for unsupervised constituency parsing \citep{spitkovsky2013breaking,Klein2004CorpusBasedIO}, or that it's effective to prune a random subset of possible trees \citep{bod2006all}. These approaches aren't necessarily orthogonal to DIORA. For instance, our model may benefit when combined with an unsupervised dependency parser.

\section{Conclusion}
\label{sec:conclusion}
In this work we presented DIORA, an unsupervised method for inducing syntactic trees and representations of constituent spans. We showed inside-outside representations constructed with a latent tree chart parser and trained with an autoencoder language modeling objective learns syntactic structure of language effectively. In experiments on unsupervised parsing, chunking, and phrase representations we show our model is comparable to or outperforms previous methods, achieving the state-of-the-art performance on unsupervised unlabeled constituency parsing for the full WSJ (with punctuation), WSJ-40, and NLI datasets. We also show our model obtains higher segment recall than a comparable model and outperforms strong baselines on phrase representations on a chunking dataset.

While the current model seems to focus primarily on syntax, future work can improve the model's ability to capture fine-grained semantics. Potential avenues include training larger models over much larger corpora, extra unsupervised or weakly-supervised phrase classification objectives, and other modeling enhancements. We are also eager to apply DIORA to other domains and languages which do not have rich linguistically annotated training sets.

\section*{Acknowledgements}

We are grateful to Carolyn Anderson, Adina Williams, Phu Mon Htut, and our colleagues at UMass for help and advice, and to the UMass NLP reading group and the anonymous reviewers for feedback on drafts of this work.
This work was supported in part by the Center for Intelligent Information Retrieval, in part by the National Science Foundation (NSF) grant numbers DMR-1534431, IIS-1514053 and CNS-0958392.
Any opinions, findings and conclusions or recommendations expressed in this material are those of the authors and do not necessarily reflect those of the sponsor.

\bibliography{naaclhlt2019}
\bibliographystyle{acl_natbib}

\clearpage

\appendix

\section{Appendices}
\label{sec:appendix}

\subsection{Composition and Input Transform \label{sec:app_composition}}

\textbf{TreeLSTM}. The $\mathrm{TreeLSTM}$ \citep{tai2015improved} function produces a hidden state vector $h$ and cell state vector $c$ given two input vectors $h_i$ and $h_j$.

\vspace{-2mm}
\begin{align*}
    \begin{bmatrix}
        x \\
        f_i \\
        f_j \\
        o \\
        u \end{bmatrix} &= 
        \begin{bmatrix}
        \sigma \\
        \sigma \\
        \sigma \\
        \sigma \\
        \tanh \end{bmatrix}
        \Bigg(
        U
    \begin{bmatrix}
        h_i \\
        h_j \end{bmatrix} + b +
        \begin{bmatrix}
        0 \\
        \omega \\
        \omega \\
        0 \\
        0 \end{bmatrix}
        \Bigg) \\
c &= c_i \odot f_i + c_j \odot f_j + x \odot u \\
h &= o + \tanh(c)
\end{align*}

The constant $\omega$ is set to $1$ for the inside, $0$ for the outside. $U$ and $b$ are learned.

\vspace{2mm}
\noindent
\textbf{MLP}. MLP (Multi-Layer Perceptron) is a deep non-linear composition with the following form:

\vspace{-2mm}
\begin{align*}
    h &= W_1 ~(W_0 ~\langle h_i, h_j \rangle + b) + b_1
\end{align*}

The operator $\langle h_i, h_j \rangle$ is a concatenation $[h_i; h_j]$. For the MLP$^{Kernel}$ $\langle h_i, h_j \rangle$ is more involved to support further interaction between the two input vectors $[h_i; h_j; h_i \odot h_j; h_i - h_j]$. The variables $W_0, W_1, b, b_1$ are learned and $c$ is unused.

\subsection{Training Details \label{sec:app_training}}

\renewcommand\labelitemi{{\boldmath$\cdot$}}


\noindent
\textit{Training Data}. Sentences of length $\leq 20$ from the SNLI and MultiNLI training sets.

\noindent
\textit{Optimization}. We train our model using stochastic gradient descent with the Adam optimization algorithm \citep{journals/corr/KingmaB14}. Cells were normalized to have magnitude of 1, following \citet{socher2011dynamic}. For instance, $\bar{a}(k) \coloneqq \bar{a}(k) / \left\Vert \bar{a}(k) \right\Vert^2$. Gradients are clipped to a maximum L2-norm of 5.

\noindent
\textit{Hyperparameters}. Chosen using grid search over cell-dimension $\{400\mathrm{D}, 800\mathrm{D}\}$ and learning rate $\{2,4,8,10,20\} \cdot 10^{-4}$.

\noindent
\textit{Early Stopping}. Using unlabeled parsing F1 against the binarized WSJ validation set.

\noindent
\textit{Vocabulary}. The model is trained in an open-vocabulary setting using pre-trained context-insensitive character embeddings from ELMo \citep{Peters:2018}.

\noindent
\textit{Batching}. Batches were constructed such that they contained sentences of uniform length. Using batch size 128 for 400D and 64 for 800D.

\noindent
\textit{Sampling}. $N$ negatives are sampled for each batch. All experiments use $N=100$.

\noindent
\textit{Training Steps}. 1M parameter updates, taking 3 days using 4x Nvidia 1080ti.

\subsection{Runtime Complexity \label{sec:app_runtime}}

The runtime complexities for DIORA's methods are shown in Table \ref{tab:runtime}. The parallel column represents the complexity when the values for all constituent pairs are computed simultaneously, assuming that these computations are independent and do not depend on values that have yet to be computed. Linear complexity is theoretically feasible depending on batch size, input length, and number of computational cores. In practice, one might experience super-linear performance.

Although both the inside pass and outside pass have an upper bound of $n^3$ operations, the outside pass will have more operations than the inside pass for sentences of length $> 1$.

As a point of reference, our implementation computes the loss over the entire WSJ corpus in 5 minutes 30 seconds at a rate of 3,500 words per second using a single GPU.

\begin{table}[ht] 
\centering
{\renewcommand{\arraystretch}{1.2} 
\begin{tabular}{l|cc} 
\toprule
Method &  Serial & Parallel \\
 \midrule
 
Inside Pass  & $O(n^3)$   & $O(n)$ \\
Outside Pass  & $O(n^3)$  & $O(n)$ \\
Training Objective  & $O(n \cdot N)$   & $O(n)$ \\
CKY   & $O(n^3)$  & $O(n)$  \\
\bottomrule

\end{tabular}
}
\caption{Runtime complexity for methods associated with DIORA in terms of sentence length $n$ and number of negative examples per token $N$. Each column represents the complexity when the values for each constituent are computed serially or in parallel.}
\label{tab:runtime}
\vspace{-4mm}
\end{table}

\subsection{Reproducing Parsing Results \label{sec:datasets}}

In Table \ref{tab:reproduce}, we've organized a reference for creating various splits of the WSJ for the purpose of evaluating unsupervised parsing.
Some splits use only the test set (section 23), others use all of the training, validation, and test data. Optionally, punctuation is stripped and sentences greater than a specified length are ignored. Predictions can be compared to the full parse trees in the annotated data, or to a binarized version. The PARSEVAL specification calculated bracketing F1 considering all spans, although some previous work diverts from PARSEVAL and ignores spans that are trivially correct (ones over the entire sentence).

\begin{table}[h] 
\centering
{\renewcommand{\arraystretch}{1.2} 
\begin{tabular}{l|ccc} 
\toprule
& WSJ & WSJ-10 & WSJ-40 \\
\midrule
Split & Test & All & Test \\
w/ Punctuation & Yes & No & No \\
Max Length & $\infty$ & 10 & 40 \\
Binarized & Yes & No & No \\
Trivial Spans & Yes & No & No \\
\bottomrule
\hline

\end{tabular}
}
\caption{Settings for unlabeled binary bracketing evaluation for different splits of the WSJ corpus.}
\label{tab:reproduce}
\end{table}

\subsection{Parse Trees}

\begin{figure}[hb]
    \centering
    \includegraphics[width=\linewidth]{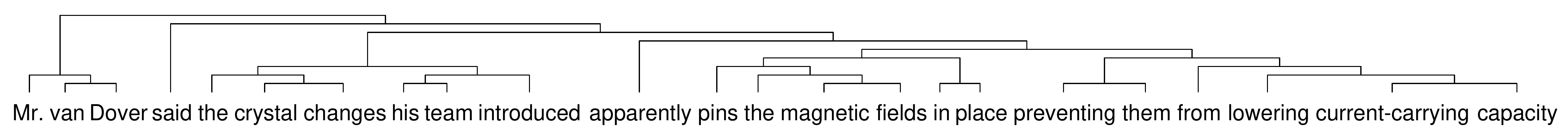}
    \includegraphics[width=\linewidth]{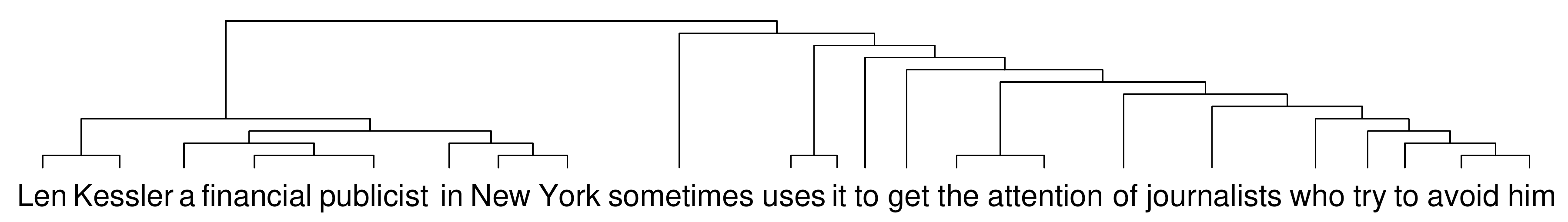}
    \includegraphics[width=\linewidth]{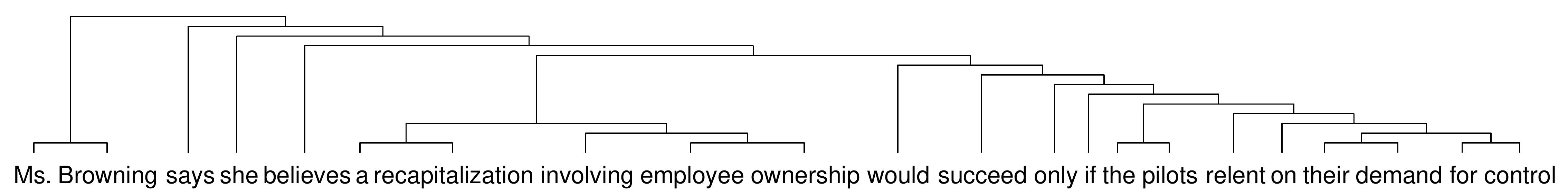}
    \includegraphics[width=\linewidth]{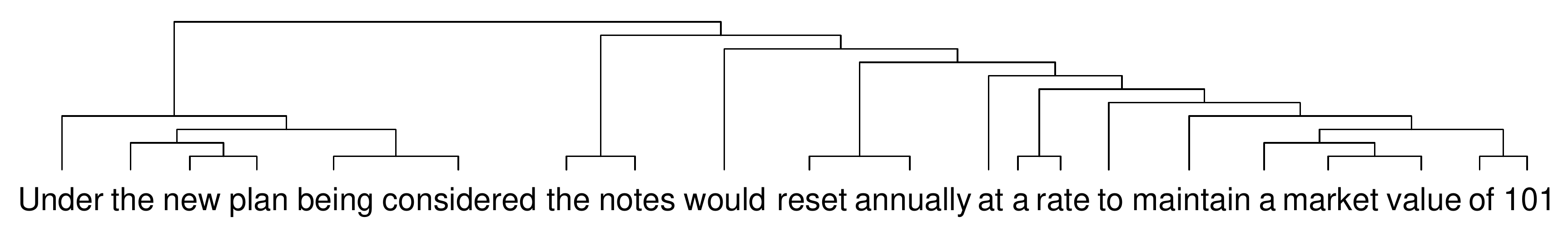}
    \includegraphics[width=\linewidth]{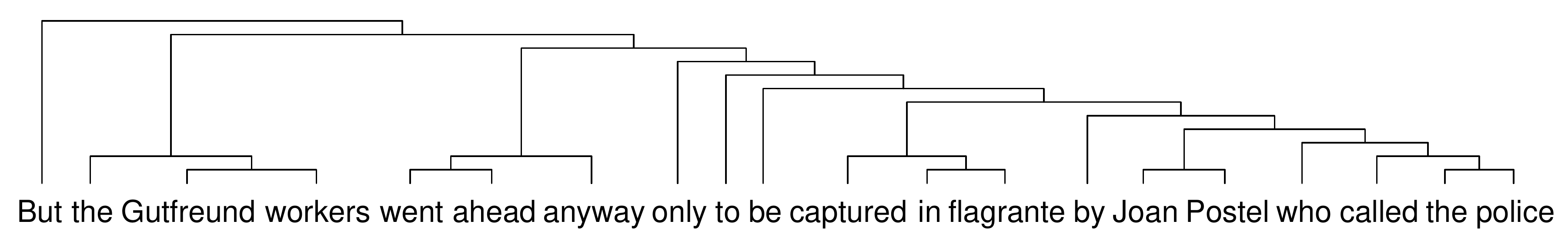}
    \includegraphics[width=\linewidth]{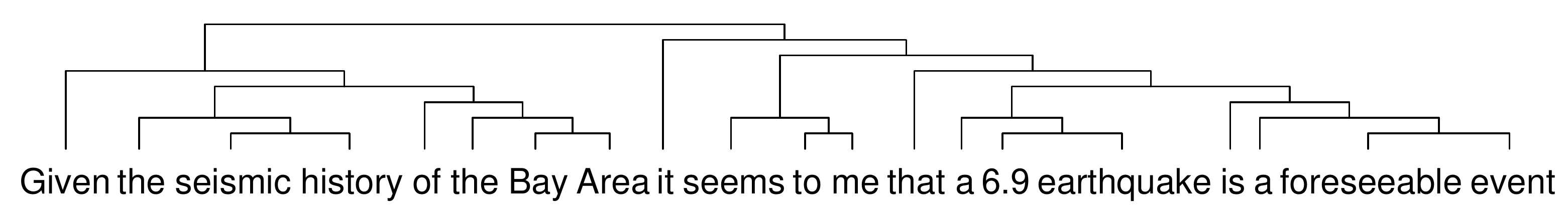}
    \includegraphics[width=\linewidth]{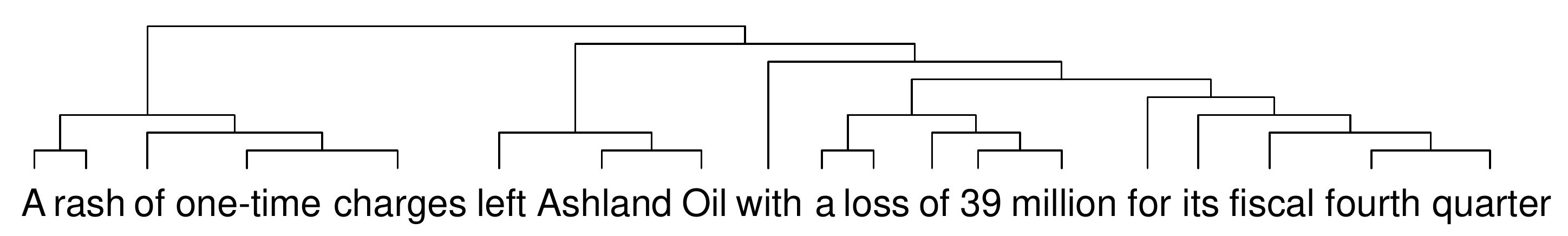}
    \includegraphics[width=\linewidth]{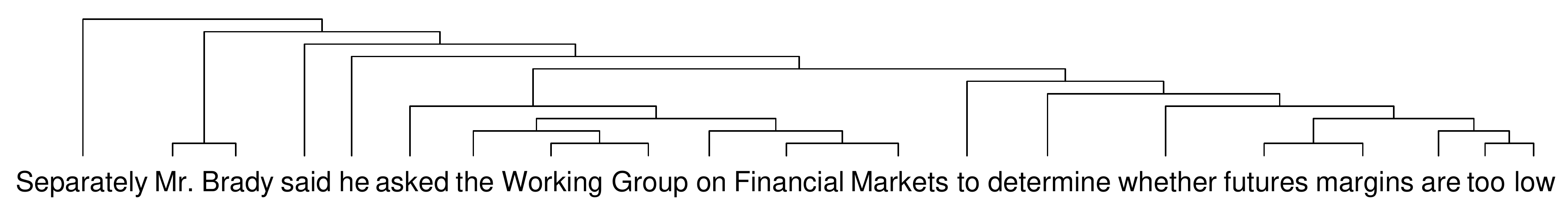}
    \includegraphics[width=\linewidth]{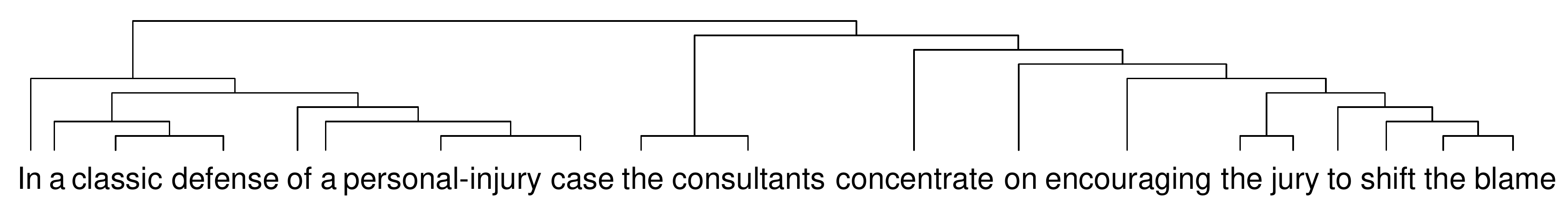}
    \includegraphics[width=\linewidth]{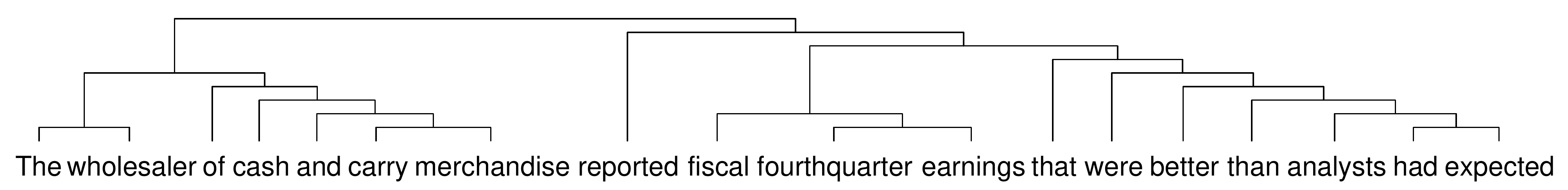}
    
    \caption{Examples where DIORA achieves 100\% accuracy compared with the binarized ground truth.}
    \label{fig:app_parses_exact_binary}
\end{figure}

\begin{figure}[h]
    \centering
    
    \includegraphics[width=\linewidth]{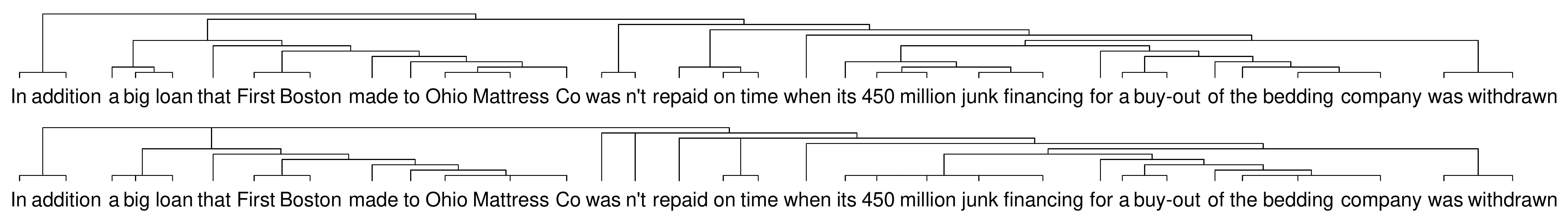}
    \includegraphics[width=\linewidth]{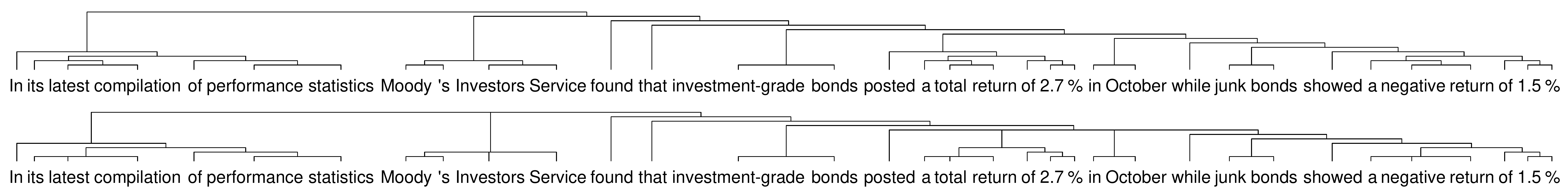}
    \includegraphics[width=\linewidth]{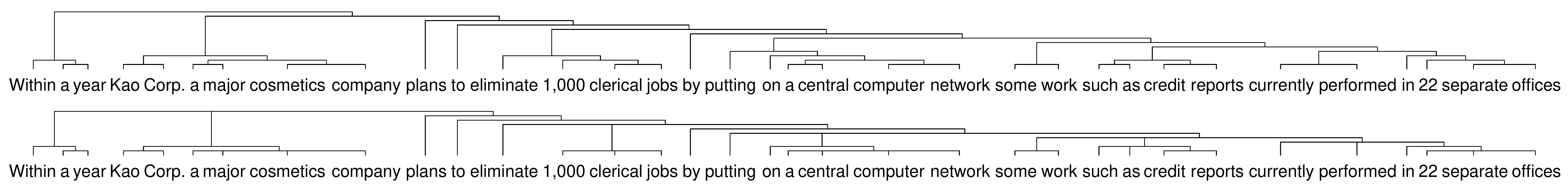}
    \includegraphics[width=\linewidth]{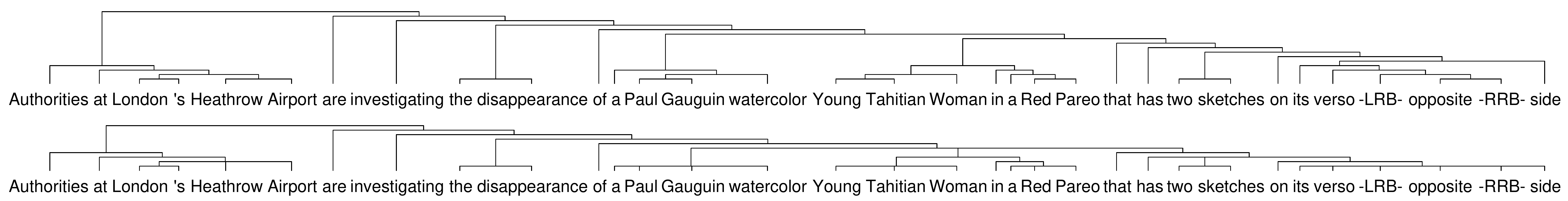}
    \includegraphics[width=\linewidth]{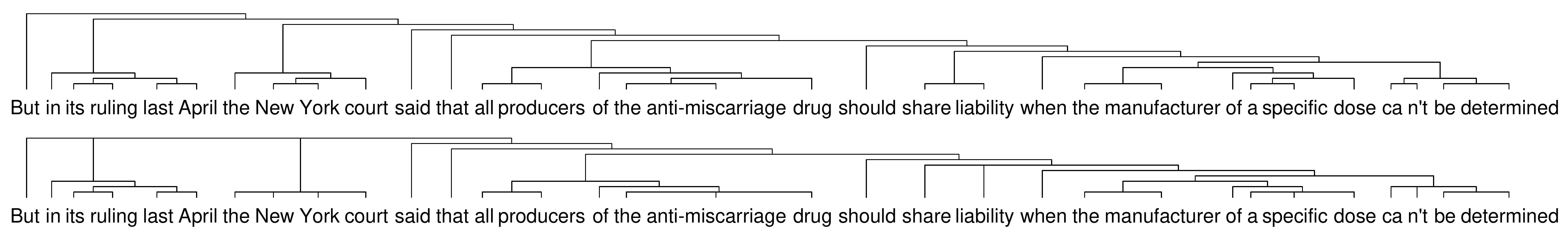}
    \includegraphics[width=\linewidth]{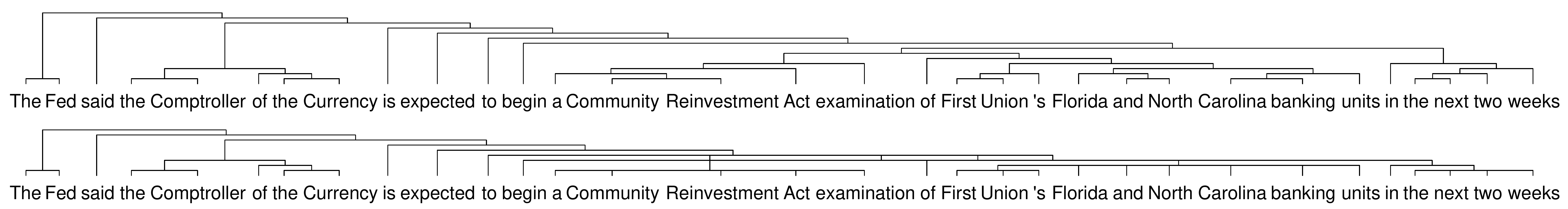}
    \includegraphics[width=\linewidth]{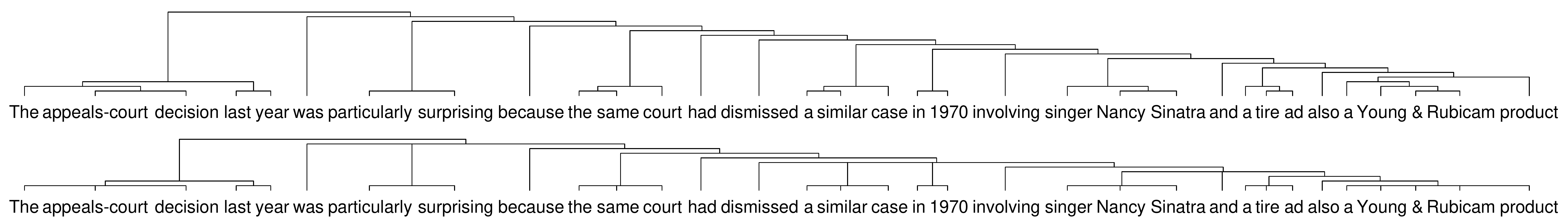}
    \includegraphics[width=\linewidth]{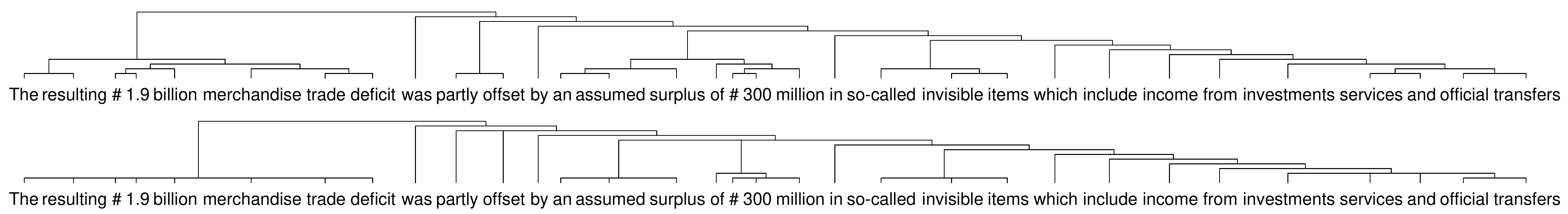}
    \includegraphics[width=\linewidth]{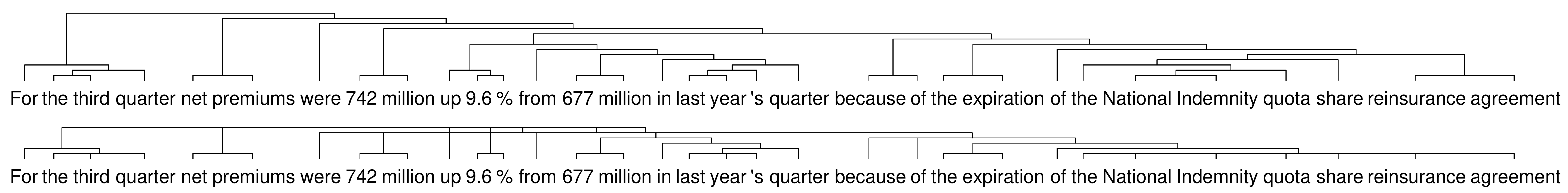}
    \includegraphics[width=\linewidth]{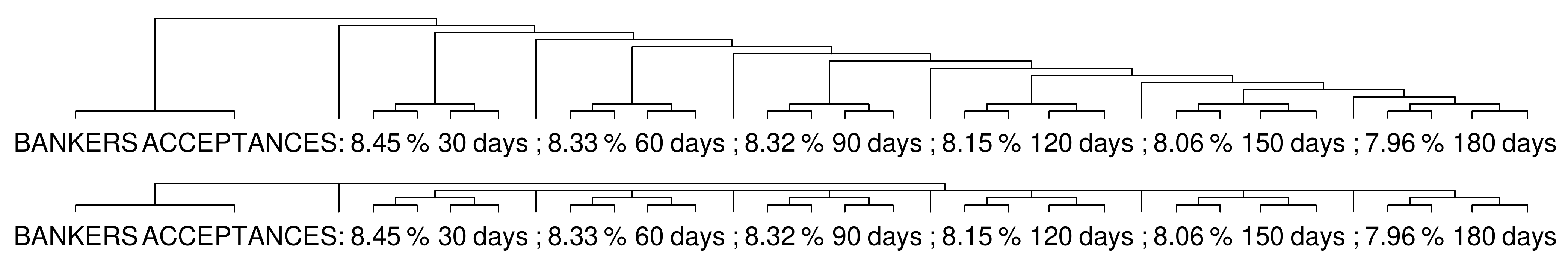}
    
    \includegraphics[width=\linewidth]{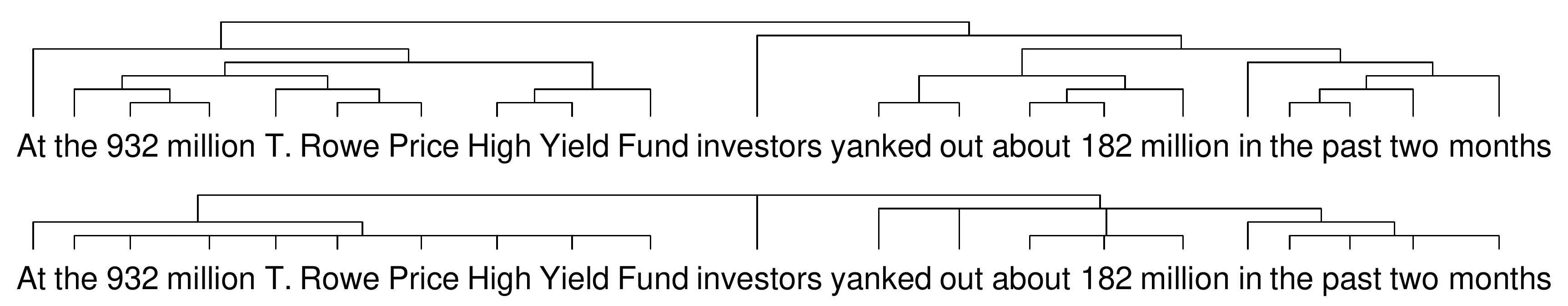}
    \includegraphics[width=\linewidth]{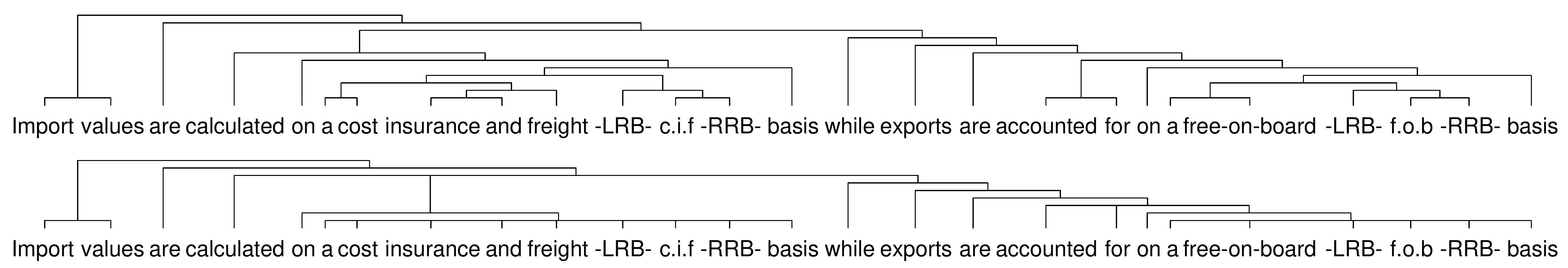}
    \includegraphics[width=\linewidth]{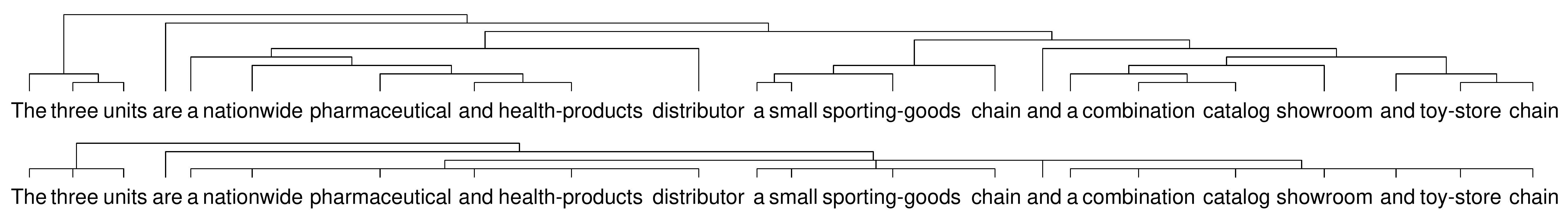}
    \includegraphics[width=\linewidth]{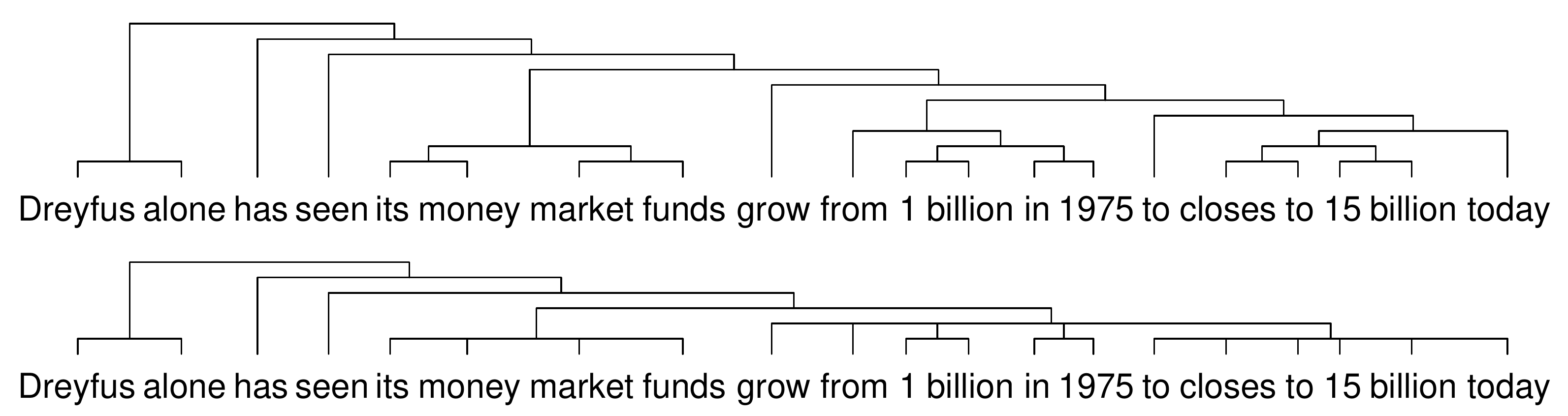}
    \includegraphics[width=\linewidth]{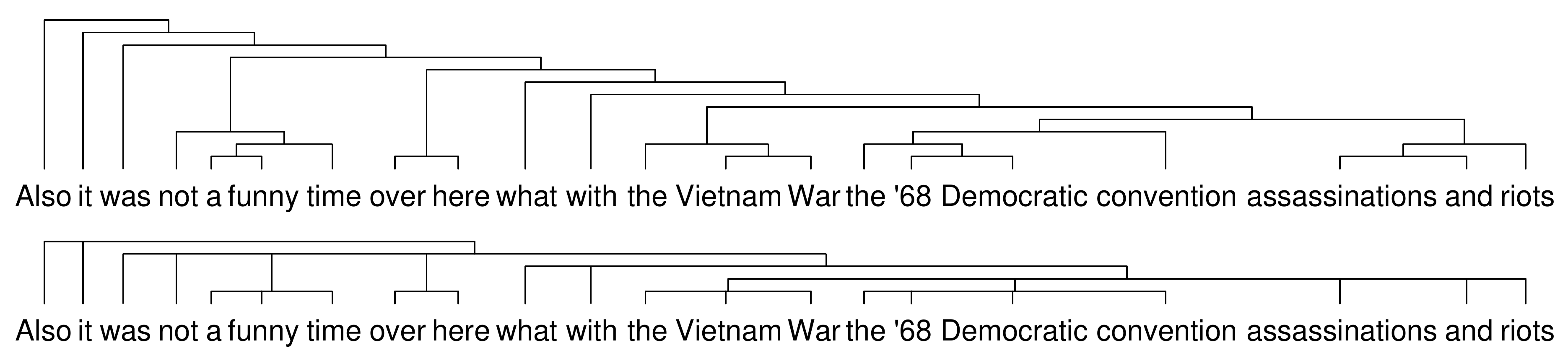}
    \includegraphics[width=\linewidth]{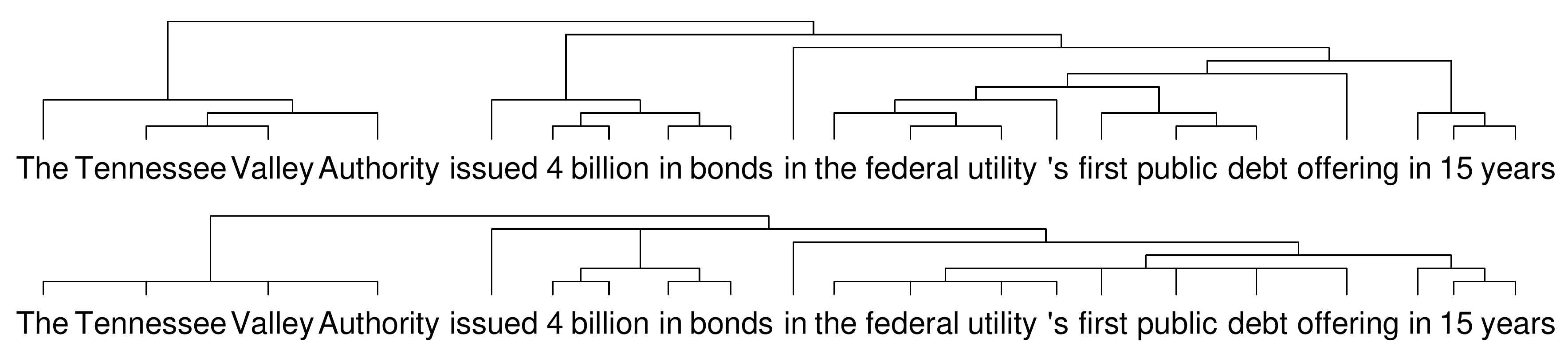}
    
    \caption{Examples where DIORA achieves 100\% recall compared with the raw (n-ary) ground truth, but less than 100\% accuracy on the binarized ground truth. DIORA is shown above the ground truth. DIORA's output is shown above the ground truth.}
    \label{fig:app_parses_exact_nary}
\end{figure}

\begin{figure*}
    \centering
    
     \includegraphics[width=\linewidth]{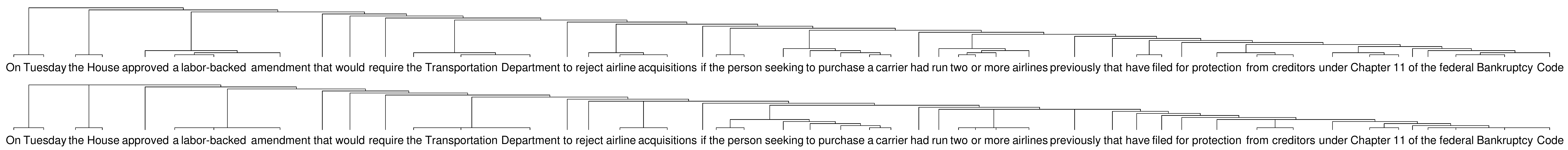} \vspace{0mm}
     
 \includegraphics[width=\linewidth]{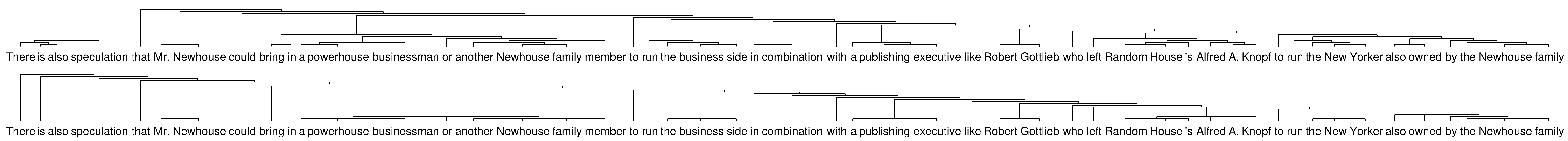} \vspace{0mm}
 
 \includegraphics[width=\linewidth]{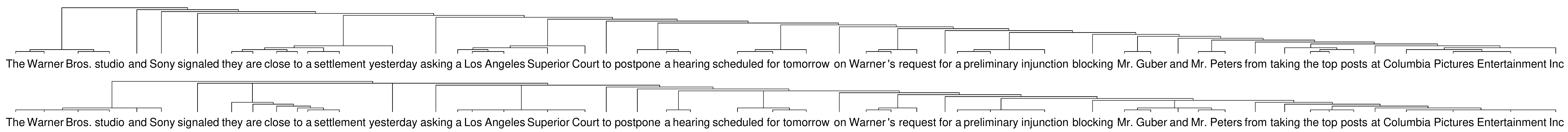} \vspace{0mm}
 
 \includegraphics[width=\linewidth]{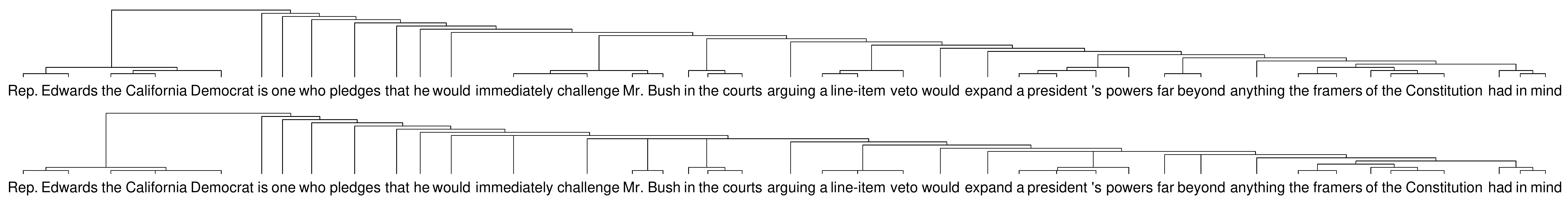} \vspace{0mm}
 
 \includegraphics[width=\linewidth]{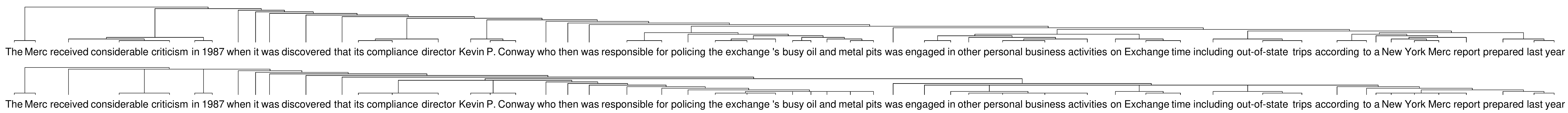} \vspace{0mm}
 
 \includegraphics[width=\linewidth]{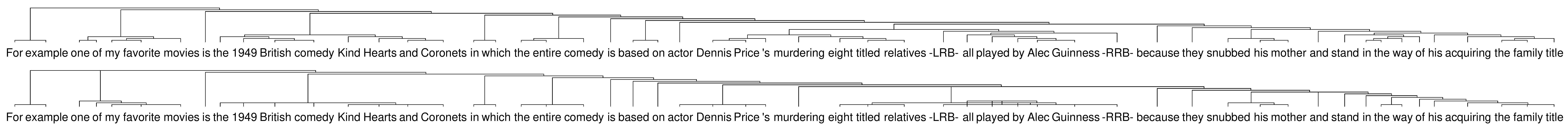} \vspace{0mm}
 
 \includegraphics[width=\linewidth]{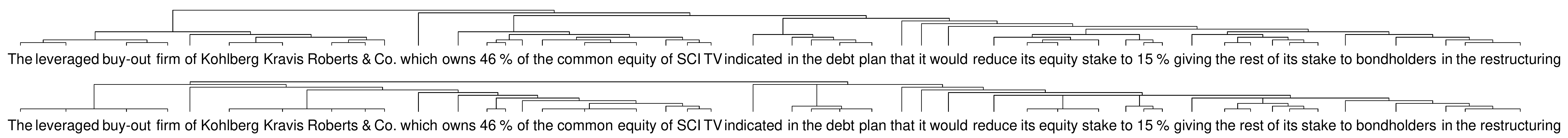} \vspace{0mm}
 
 \includegraphics[width=\linewidth]{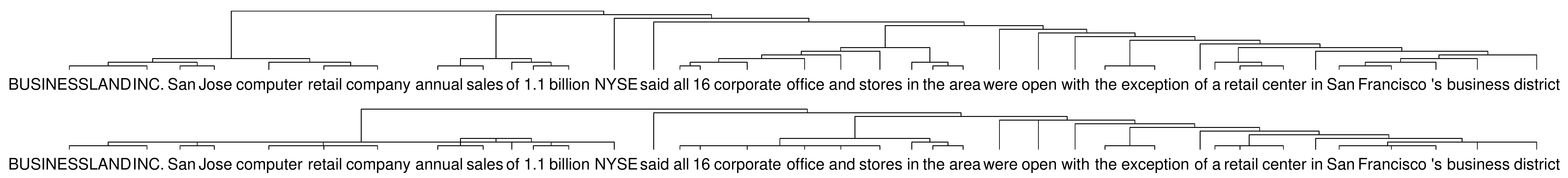} \vspace{0mm}
 
 \includegraphics[width=\linewidth]{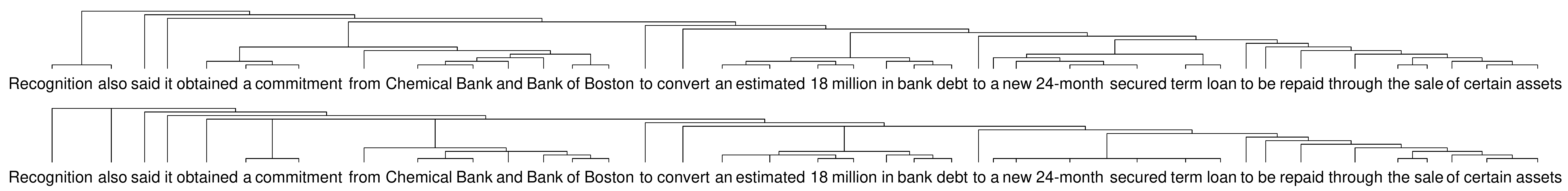} \vspace{0mm}
 
 \includegraphics[width=\linewidth]{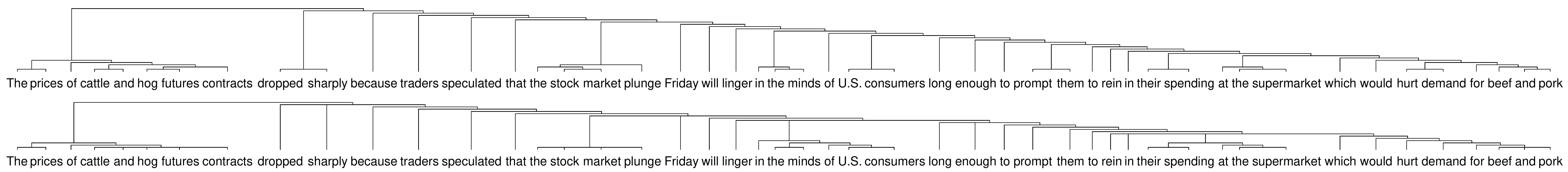}
    
    \caption{DIORA can perform close to the ground truth even on long sentences. 
    In this figure, n-ary trees are shown for the ground truth. DIORA's output is shown above the ground truth.}
    
    \label{fig:app_parses_extra}
\end{figure*}

Examples of parse trees derived from the compatibility scores are shown in Figures \ref{fig:app_parses_exact_binary}, \ref{fig:app_parses_exact_nary}, and \ref{fig:app_parses_extra}. Some punctuation has been removed for easier readability.

\newpage

\newpage

\end{document}